\documentclass[runningheads]{llncs}
\usepackage{graphicx}

\usepackage{tikz}
\usepackage{comment}
\usepackage{amsmath,amssymb} 
\usepackage{color}
\usepackage{graphicx}
\usepackage{booktabs}
\usepackage{multirow}
\usepackage{array}
\usepackage{graphicx}
\usepackage{wrapfig}
\usepackage{subcaption}

\usepackage{enumitem}
\usepackage[accsupp]{axessibility}  

\usepackage[width=122mm,left=12mm,paperwidth=146mm,height=193mm,top=12mm,paperheight=217mm]{geometry}

\begin{document}
\pagestyle{headings}
\mainmatter

\title{Planes vs. Chairs: Category-guided 3D shape learning without any 3D cues} 

\titlerunning{Planes vs. Chairs: Category-guided 3D shape learning without any 3D cues} 
\authorrunning{Z. Huang et al.} 
\author{Zixuan Huang\inst{1} \and
Stefan Stojanov\inst{1} \and
Anh Thai\inst{1} \and \\
Varun Jampani\inst{2} \and
James M. Rehg\inst{1}}
\institute{Georgia Institute of Technology \and Google Research}

\maketitle

\begin{abstract}
  We present a novel 3D shape reconstruction method which learns to predict an implicit 3D shape representation from a single RGB image. Our approach uses a set of single-view images of multiple object categories without viewpoint annotation, forcing the model to learn across multiple object categories without 3D supervision. To facilitate learning with such minimal supervision, we use category labels to guide shape learning with a novel categorical metric learning approach. We also utilize adversarial and viewpoint regularization techniques to further disentangle the effects of viewpoint and shape. We obtain the first results for large-scale (more than 50 categories) single-viewpoint shape prediction using a single model without any 3D cues. We are also the first to examine and quantify the benefit of class information in single-view supervised 3D shape reconstruction. Our method achieves superior performance over state-of-the-art methods on ShapeNet-13, ShapeNet-55 and Pascal3D+.


\end{abstract}

\section{Introduction}
\label{section:introduction}
\begin{figure*}[t]
    \centering
    \includegraphics[width=1\linewidth]{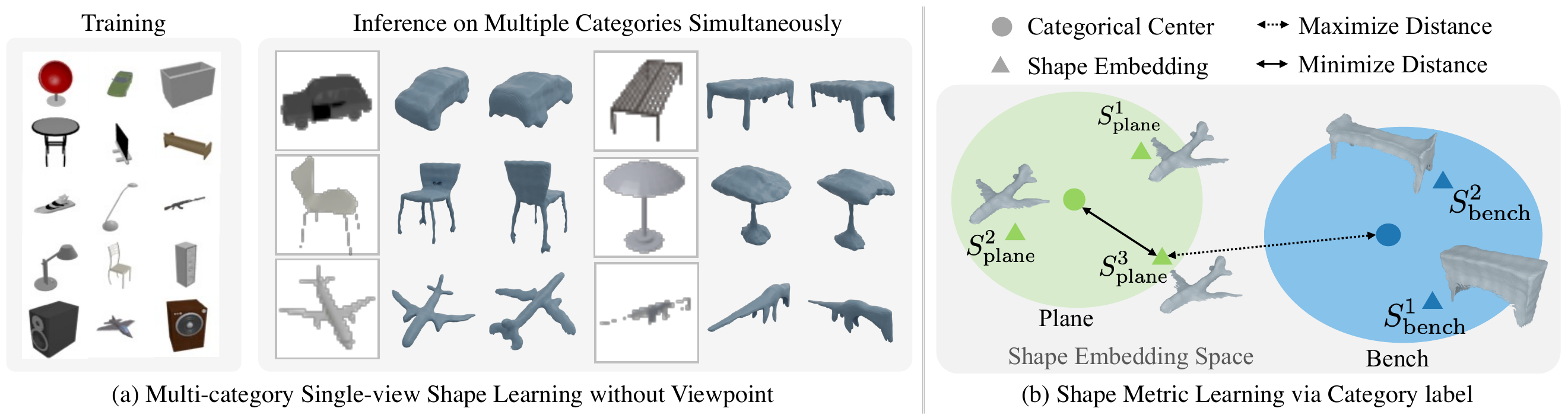}
    \vspace{-20pt}
    \caption{(a) We present the first method to learn 3D shape reconstruction from single-viewpoint images over \emph{multiple object categories simultaneously}, without 3D supervision or viewpoint annotations. (b) To facilitate shape learning under such a challenging scenario, We leverage category labels to perform metric learning in the shape embedding space. For each category, we learn a shape center. Then with a given sample, we minimize the distance between its shape embedding and the corresponding category center, while contrasting this distance with other inter-category distances.}
    \label{fig:teasor}
    \vspace{-20pt}
\end{figure*}

Reconstructing the 3D shape of objects from 2D images is a fundamental computer vision problem. 
Recent works have demonstrated that high-quality
3D shape reconstructions can be obtained from a single RGB input  image~\cite{choy20163d,Gkioxari_2019_ICCV,groueix2018,mescheder2019occupancy}.
In the most straight-forward approach, explicit 3D supervision from ground-truth (GT) object shapes is used to train a shape reconstruction model~\cite{thai20203d,mescheder2019occupancy,saito2019pifu,xu2019disn}. While this approach yields strong performance, GT 3D shape is difficult to obtain on a large scale.
This limitation can be addressed via multi-view supervision~\cite{yan2016perspective,tulsiani2017multi,yang2018learning,lin2018learning,insafutdinov2018unsupervised,kato2018neural,liu2019soft,niemeyer2020differentiable}, in which the learner optimizes a reprojection loss over a set of viewpoints. This approach becomes more difficult when fewer viewpoints are available during training. 
The limiting case is single-view supervision, which is challenging due to the well-known entanglement between shape and pose.\footnote{When only a single view of each object instance is available for training, there are an infinite number of possible 3D shapes that could explain the image under appropriate camera viewpoints.}
Prior works (see Table~\ref{tab:method-comparison}) have addressed this challenge in two ways. The first approach assumes that the camera viewpoint is known at training time~\cite{lin2020sdf,kato2019learning}. While this strong assumption effectively reduces entanglement, it is not scalable as it requires pose annotations. 

A second line of attack trains separate models for each category of 3D objects~\cite{kanazawa2018learning,tulsiani2020implicit,goel2020shape,li2020self,henzler2019escaping,henderson2019learning,ye2021shelf,navaneet2020image,wu2020unsupervised}. This single category approach provides the learner with a strong constraint -- each model only needs to learn the ``pattern" of shape and pose entanglement associated with a single object category. This approach has the disadvantage that data cannot be pooled across categories: pooling data can be beneficial to tasks such as generalization to unseen categories of objects (zero-shot~\cite{kar2017learning,zhang2018learning} or few-shot~\cite{yang2018learning}) and viewpoint learning. For example, tables and chairs have many features (e.g. legs) in common, and a multi-class reconstruction model could leverage such similarities in constructing shared feature representations.
We introduce a novel \emph{multi-category, single-view (MCSV)} 3D shape learning method, shown in Fig.~\ref{fig:teasor} (a), that does not use shape or
viewpoint supervision, and can learn to represent multiple object categories within a single reconstruction model.
Our method exploits the observation that shapes within a category (e.g. planes) 
are more similar to each other than they are to the shapes in other categories (e.g. planes vs. chairs). We use a shape metric learning approach, illustrated in Fig.~\ref{fig:teasor}~(b), which simultaneously learns category shape embeddings (category centroids denoted as green/blue dots) together with instance shape embeddings (green/blue triangles) and minimizes the distance between shape instances and their corresponding category centers (solid arrow) while maximizing the distance to the other categories (dotted arrow). 
This method has two benefits. First, the learned shape manifold captures a strong prior knowledge about shape distance, which helps to eliminate erroneous shapes that might otherwise explain the input image. Second, similar shapes tend to be clustered together, enabling the supervision received by a particular shape to affect its neighbors, and helping to overcome the limitation of having only a single viewpoint.
As a consequence of this approach, we are able to train a single model on all 55 object categories in ShapeNet-55, which has not been accomplished in prior work.

Another important aspect of our approach is that we represent object shape as an implicit signed distance field (SDF) instead of the more commonly used mesh representation, in order to accommodate objects with varying topologies. This is an important goal in scaling MCSV models to handle large numbers of object categories. Specifically, our model (Fig.~\ref{fig:teasor} (a)) takes a single input image and produces an implicit SDF function for the 3D object shape. In contrast to mesh and voxel representations, SDF representations create challenges in training because they are less constrained and require a more complex differentiable rendering pipeline. A beneficial aspect of our approach is the use of adversarial learning~\cite{henzler2019escaping,ye2021shelf} 
and cycle-consistency~\cite{mustikovela2020self,ye2021shelf} for regularization.


In summary, our 3D reconstruction learning approach has the following favorable properties:
\begin{itemize}[leftmargin=*]
\setlength\itemsep{-0.05em}
\item \textbf{Extremely weak supervision.} We use only single-view masked images with 
category labels during training (considered as `self-supervised' in prior works).
\item \textbf{Categorical shape metric learning.} We make effective use of class labels with our novel
shape metric learning.
\item \textbf{Single template-free multi-category model.} 
We learn a single model to reconstruct multiple object categories without using any category-specific templates.
\item \textbf{Implicit SDF representation.} We use implicit SDFs, instead of meshes, to represent diverse object topologies.
\item \textbf{Order of magnitude increase in object categories learned in a single model.}
Due to the scalability of our approach, we are the first to train a MCSV model on all 55 categories of ShapetNetCoreV2~\cite{chang2015shapenet}.
\item \textbf{State-of-the-art reconstructions.} Experiments on both ShapeNet~\cite{chang2015shapenet} and 
Pascal3D+~\cite{xiang2014beyond} demonstrate consistently better performance than previous methods.
\end{itemize}

\section{Related Work}
\label{section:related_work}

\vspace{1mm} \noindent \textbf{Single-View Supervision.} The body of work that is most closely-related to this paper are methods that use single-view supervision to learn 3D shape reconstruction with a back-projection loss~\cite{kanazawa2018learning,tulsiani2020implicit,goel2020shape,li2020self,henzler2019escaping,henderson2019learning,gadelha20173d,ye2021shelf,navaneet2020image,kato2019learning,wu2020unsupervised,lin2020sdf}. These works can be organized as depicted in Table~\ref{tab:method-comparison} and largely differ in their choice of 1) learning a single multi-category (multi-class) vs multiple single-category models for reconstruction; 2) shape representation (e.g. implicit SDF vs explicit mesh); 3) known vs. unknown viewpoint assumption. We are the first to demonstrate the feasibility of single-view, multi-category learning of an implicit shape representation (SDF) under the unknown viewpoint condition via the use of category labels. We are also the first to provide results on all of the ShapeNet-55 classes in a single model, an order of magnitude more than prior works. 

\begin{table}
    \vspace{-0pt}
    \centering
    \caption{Single-view supervised methods for shape reconstruction. For methods without supervision, they are mainly verified under a limited setting~\cite{henderson2019learning} or for specific objects~\cite{wu2020unsupervised}. K: keypoints, T: category templates, M: masks, m: mesh, vox: voxel, pc: pointcloud, im: implicit representation.}
    \begin{tabular}{|c|c|c|c|c|c|c|c|c|c|c|c|c|c|c|}  \hline
    Model & \cite{kanazawa2018learning} & \cite{tulsiani2020implicit} & \cite{goel2020shape} & \cite{li2020self} & \cite{gadelha20173d} & \cite{henzler2019escaping} & \cite{henderson2019learning} & \cite{ye2021shelf} & \cite{navaneet2020image} & \cite{lin2020sdf} & \cite{wu2020unsupervised}   & \cite{kato2019learning} & \cite{simoni2021multi} & Ours \\ \hline 
    Multi-Class Rec. & - & - & - & - & - & - & - & - & - & - & - & \checkmark & \checkmark & \checkmark \\
    Supervision & K,M & M,T & M,T & M & M & M & - & M & M & M,V & - & M,V  & M,V & M \\
    3D Rep. & m & m & m & m & vox & vox & m & m & pc & im & d & m & m & im \\
    \hline
    \end{tabular}
    \label{tab:method-comparison}
    \vspace{-15pt}
\end{table}

Within this body of work~\cite{kato2019learning,lin2020sdf,ye2021shelf} and the concurrent work of~\cite{simoni2021multi} are the most closely related. Kato et. al.~\cite{kato2019learning} (VPL) and Simoni et. al.~\cite{simoni2021multi} are the only prior works to address multi-category shape learning with a single model, but both assume access to ground truth (GT) viewpoint and use mesh representations. Similar to our work, Ye et. al.~\cite{ye2021shelf} do not use viewpoint GT,
but they train one model for each object category. Further, their mesh prediction is refined from a low-resolution voxel prediction by optimizing on each individual sample. The latter is less efficient than feedforward models and can potentially harm the prediction of concave shapes or details. In contrast with~\cite{kato2018neural,simoni2021multi,ye2021shelf} we use an implicit shape representation. Further, as shown in Table~\ref{tab:method-comparison}, some earlier approaches use fixed meshes as deformable shape templates. In contrast with implicit functions, meshes with fixed topology restrict the possible 3D shapes that can be accurately represented. 


Lin et. al.~\cite{lin2020sdf} is the only prior work to use an implicit SDF representation for single view supervision, and we adopt their SDF-SRN network architecture and associated reprojection losses in our formulation. This work trains single-category models and assumes that the camera viewpoint is known, whereas one of the main goals of our work is to remove these assumptions and to demonstrate that the unconstrained SDF representation can be successful in the multi-category, unknown viewpoint case. We consistently outperform a version of SDF-SRN modified for our setting in Table~\ref{tbl:main-shapenet13}, which highlights our innovations.

\vspace{1mm} \noindent \textbf{Shape Supervision.} 
Prior works on single image 3D reconstruction with explicit 3D geometric supervision have achieved impressive results~\cite{choy20163d,Gkioxari_2019_ICCV,groueix2018,mescheder2019occupancy,xu2019disn}. However, the requirement of 3D supervision limits the applicability of these methods. To overcome this, subsequent works employ multi-view images for supervision. The use of image-only supervision is enabled by differentiable rendering, which enables the generation of 2D images/masks from the predicted 3D shape and the comparison to the ground truth reference images as supervision. These methods can be grouped by their representation of shape, including voxels~\cite{yan2016perspective,tulsiani2017multi,yang2018learning}, pointclouds~\cite{lin2018learning,insafutdinov2018unsupervised}, meshes~\cite{kato2018neural,liu2019soft} and implicit representation~\cite{niemeyer2020differentiable}. In contrast to these works, we allow only  \emph{a single image per object instance} in the training dataset.

\vspace{1mm} \noindent \textbf{Category Information.} 
Few prior works have explored using category-specific priors for few-shot shape reconstruction~\cite{wallace2019few,michalkiewicz2020few}, where they assume voxel templates are available for each category. Our method uses category priors for shape reconstruction, but by only leveraging the significantly weaker supervision of category labels.

\vspace{1mm} \noindent \textbf{Deep Metric Learning.}
To harness shape learning with category labels, our method also shares ideas with metric learning. The key idea of metric learning is to learn an embedding space where similar instances are together and dissimilar instances are far according to some distance metric. This paradigm has shown success in self-supervised~\cite{chen2020simple,he2020momentum} and supervised~\cite{khosla2020supervised} learning, as well as many downstream tasks e.g. person re-identification~\cite{hermans2017defense}.
We use the idea of metric learning for single-view shape reconstruction and demonstrate its effectiveness for the first time. 

\section{Approach}
\label{section:approach}

\begin{figure*}[t]
\centering
	\includegraphics[width=0.95\linewidth]{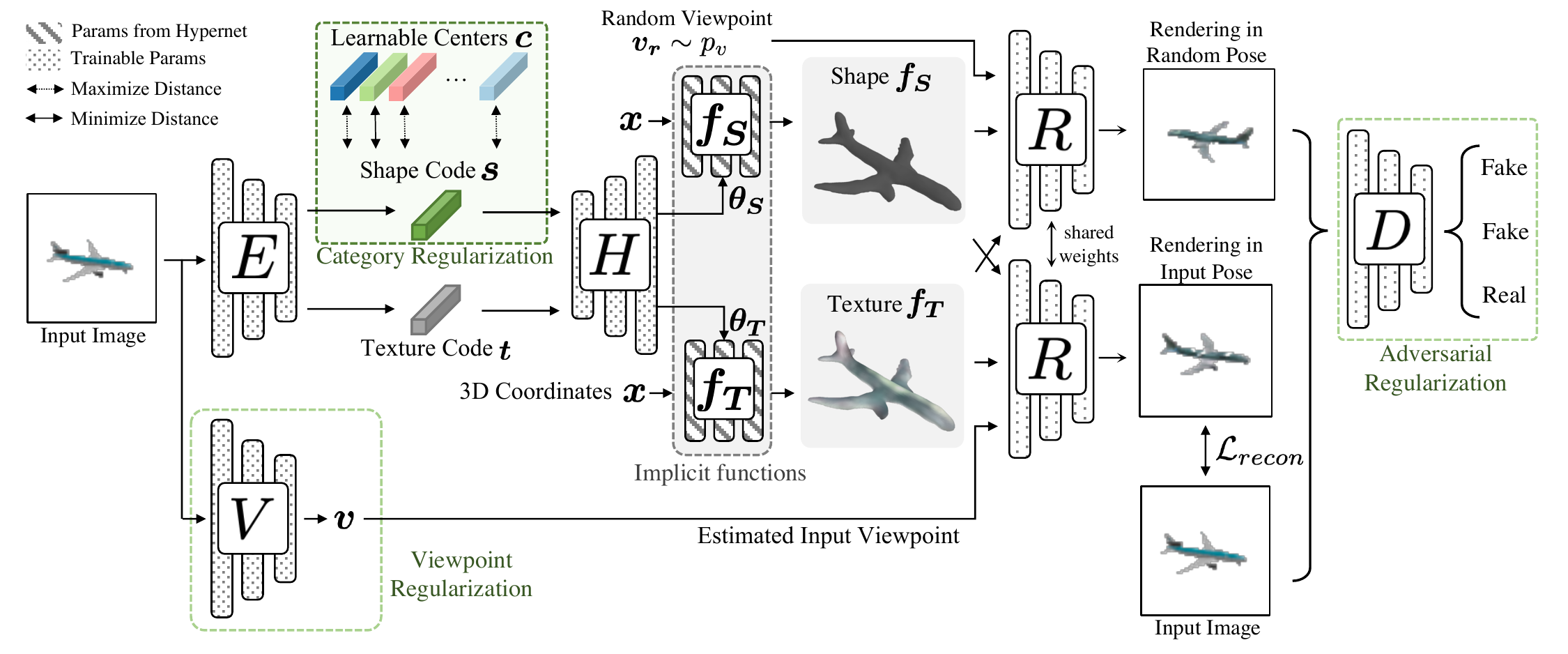}
	\vspace{-10pt}
	\caption{\textbf{Approach Overview.} Given the input image, the encoders $\boldsymbol{E}$ and $\boldsymbol{V}$ predicts the shape $\boldsymbol{s}$ and texture $\boldsymbol{t}$ codes and viewpoint $\boldsymbol{v}$. The hypernetwork $\boldsymbol{H}$ takes both latent codes and outputs the parameters $\boldsymbol{\theta_S}$, $\boldsymbol{\theta_T}$ of the implicit functions $\boldsymbol{f_S}$ and $\boldsymbol{f_T}$. These functions are sampled with learnable ray tracer $\boldsymbol{R}$ to render the image with the predicted viewpoint $\boldsymbol{v}$ and thus to compute the reconstruction loss. We also render an extra image from a random viewpoint for adversarial and viewpoint regularizations. Our shape metric learning technique is performed over the shape latent code.} \vspace{-1.5em}
	\label{fig:overview}
\end{figure*}

Given a collection of $n$ images concatenated with their masks $\{I_i\in\mathbb{R}^{h \times w \times 4}\}_{i=1}^n$ and class labels $\{\mathbf{y}_i \in \{0,1,\cdots, c\}\}$,
our goal is to learn a single-view 3D reconstruction model without any 
3D, viewpoint, or multi-view supervision.
In contrast to most existing work, we learn a single network that works
across all object categories.
We represent shape using an implicit shape representation function $\boldsymbol{f_S}: \mathbb{R}^3 \rightarrow \mathbb{R}$, a multi-layer perceptron (MLP) that maps 3D coordinates to signed distance function (SDF) values $s \in \mathbb{R}$. Following a standard framework for learning 3D shape via differentiable rendering, our model first infers shape, texture and viewpoints from the input images, which are then rendered into images.
The rendered image is then compared to the input image, 
providing a training signal for model learning. However, in our challenging multi-category setting, without viewpoint or 3D supervision, providing supervision only by comparing the rendered image and input images of the object results in poor performance. To mitigate this, we propose a set of regularization techniques which improves reconstruction quality and stabilizes model training.
We first present an overview of different modules of our approach in Section~\ref{subsection:overview}, and then introduce our category-based 
shape metric learning in Section~\ref{subsection:category}. Finally, we present other regularization methods in Section~\ref{subsection:regularization}. Implementation details and license are described in the appendix.

\subsection{Network Overview}
\label{subsection:overview}

Our model is trained in an end-to-end manner and can be decomposed into four main network modules (Fig.~\ref{fig:overview}).

\vspace{1mm} \noindent \textbf{Image encoder.} The image encoder $\boldsymbol{E}$ maps an image with mask channel $I \in \mathbb{R}^{h \times w \times 4}$ to a latent shape vector $\boldsymbol{s} \in \mathbb{R}^{l}$ for the downstream shape predictor, and a latent texture vector $\boldsymbol{t} \in \mathbb{R}^{l}$ for texture predictor. 

\vspace{1mm} \noindent \textbf{Shape and texture prediction module.} Following the design of hypernetworks~\cite{ha2016hypernetworks,lin2020sdf}, the shape and texture prediction module uses latent codes $\boldsymbol{s}$ and $\boldsymbol{t}$ to predict the parameters of shape and texture implicit functions $\boldsymbol{f_S}$ and $\boldsymbol{f_T}$ that map 3D query points $\boldsymbol{x}$ to SDF and texture predictions: 
\begin{align}
    &{\boldsymbol{\theta_S}, \boldsymbol{\theta_T}} \leftarrow \boldsymbol{H}(\boldsymbol{s}, \boldsymbol{t}; \boldsymbol{\theta_H}) \\
    &\boldsymbol{f_S}:(\boldsymbol{x}; \boldsymbol{\theta_S}) - \text{SDF prediction} \\
    &\boldsymbol{f_T}:(\boldsymbol{x}; \boldsymbol{\theta_T}) - \text{Texture prediction}
\end{align}
$\boldsymbol{f_S}$ and $\boldsymbol{f_T}$ are MLPs with predefined structure and parameters estimated by the hypernet $\boldsymbol{H}$ with parameters $\boldsymbol{\theta_H}$. 

\vspace{1mm} \noindent \textbf{Viewpoint prediction module.} In our model, the viewpoint is represented as the trigonometric functions of Euler angles for continuity~\cite{beyer2015biternion}, i.e. $\boldsymbol{v} = [\cos \boldsymbol{\gamma}, \sin \boldsymbol{\gamma}]$ with $\boldsymbol{\gamma}$ denoting 3 Euler angles. The viewpoint prediction network $\boldsymbol{V}(I;\boldsymbol{\theta_V})$ predicts the viewpoint $\boldsymbol{v} \in \mathbb{R}^6$ of the object in the input image $I$ with regard to a canonical pose (can be different from the human-defined canonical pose), where $\boldsymbol{\theta_V}$ are the learnable parameters. 

\vspace{1mm} \noindent \textbf{Differentiable renderer.} We use an SDF-based implicit differentiable renderer from~\cite{lin2020sdf} which takes the shape, texture and the viewpoint as inputs and renders the corresponding RGB image. Formally, we denote it as a functional, $\boldsymbol{R}(\boldsymbol{f_S}, \boldsymbol{f_T}, \boldsymbol{v}; \boldsymbol{\theta_R})$, which maps shape SDF, texture implicit function and the viewpoint into 2D RGB image with an alpha channel, in $\mathbb{R}^{h \times w \times 4}$. The renderer itself is also learnable with $\boldsymbol{\theta_R}$ parameters. Note that as the renderer in~\cite{lin2020sdf} cannot render masks, we modify it to render an extra alpha channel from SDF field following~\cite{yariv2020multiview}. 

Before we describe different regularizations, we briefly discuss the major challenge in learning our model. During training, the only supervision signals are the input images and the masks of the objects. The common approach is to minimize a reconstruction loss via differentiable rendering\footnote{Masks are leveraged with additional losses that enforce consistency between the shape and the input mask. For conciseness, we omit them here and refer readers to~\cite{lin2020sdf} for more details.}
\begin{equation}
    \mathcal{L}_{recon} = \|I-\boldsymbol{R}(\boldsymbol{f_S}, \boldsymbol{f_T}, \boldsymbol{v}; \boldsymbol{\theta_R})\|^2_2.
\end{equation}
This learning scheme works well with multi-view supervision~\cite{niemeyer2020differentiable}, or even single-view images with viewpoint ground truth~\cite{lin2020sdf}. However, when there is only single-view supervision without viewpoint ground truth, this is significantly more difficult. There are infinite combinations of shapes and viewpoints that perfectly render the given image, and the model does not have any guidance to identify the correct shape-viewpoint combination. To tackle this problem,
we utilize category labels to guide the learning of shape via metric learning. We also 
make use of other regularizations including adversarial learning and cycle consistency to facilitate learning.

\subsection{Shape metric learning with category guidance}
\label{subsection:category}

\begin{figure}[h]
    \begin{center}
        \includegraphics[width=0.9\linewidth]{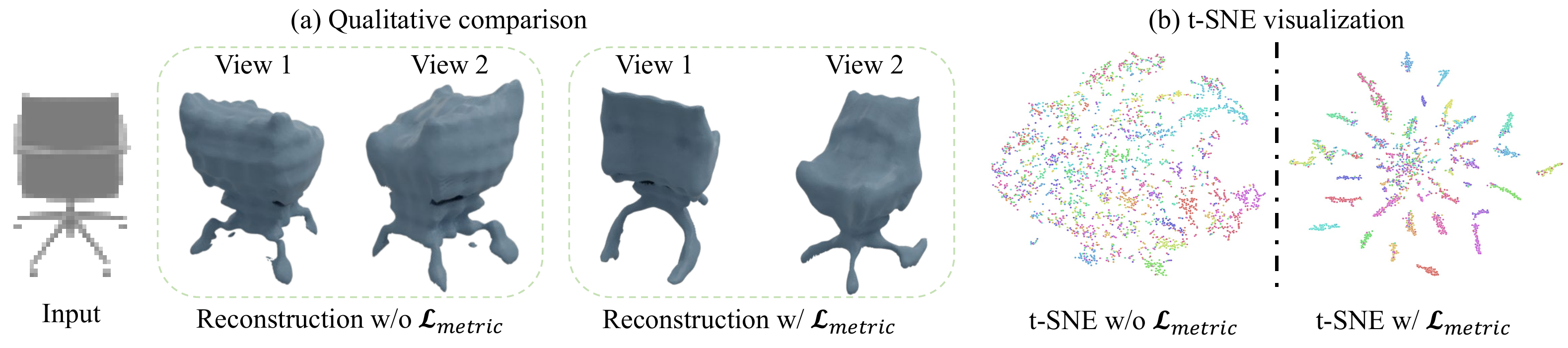}
    \end{center}
	\caption{(a) Category-guided metric learning is beneficial for eliminating erroneous shapes that can explain input image. In the first view (close to input viewpoint), both shapes looks plausible and can explain input image. But from another view, it is clear that our metric-learned shape is more reasonable as a chair, while the baseline can be closer to a table/nightstand. (b) t-SNE visualization of the shape embedding space on ShapeNet-55. Category-guided metric learning results in a more structured representation and facilitates joint learning.}
	\label{fig:metric_comp_tsne}
	\vspace{-15pt}
\end{figure}

Our novel shape metric learning approach is the key ingredient that enables us to train a single model containing 55 shape categories. The key idea is to learn a metric which maps shapes within the same category (e.g. chairs) to be close to each other, while mapping shapes in different categories (e.g. planes vs. chairs) to be farther apart. This approach leverages qualitative label information to obtain a continuous metric for shape similarity, resulting in two benefits. First, the shape manifold defined by the learned metric constrains the space of possible 3D shapes, helping to eliminate spurious shape solutions that would otherwise be consistent with the input image (see Fig.~\ref{fig:metric_comp_tsne} (a) for an example). Second, the shape manifold tends to group similar shapes together, creating a neighborhood structure over the training samples. For example, without category guidance it can be difficult for the network to learn the relationship between two images of different chairs captured from different, unknown, camera viewpoints. However, with the learned shape manifold these instances are grouped together, facilitating the sharing of supervisory signals from the rendering-based losses. A qualitative reconstruction comparison in Fig.~\ref{fig:metric_comp_tsne} (a) and t-SNE visualizations of shape embedding space in Fig.~\ref{fig:metric_comp_tsne} (b) provide further insight into these benefits. Note that as the image discriminator used for adversarial regularization uses images and labels as input, category information is still available during training even without metric learning. Therefore, the current comparison demonstrates the  effectiveness of metric learning for further utilizing category label information.

Formally, for a batch of $n$ images ${\{I_i\}}_{i=1}^n$, we extract both shape latent codes and texture latent codes via an image encoder $\boldsymbol{E}$. For the $i^{th}$ sample, we denote the latent shape code as $\boldsymbol{s}_i$ and the category label as $y_i \in \mathcal{C}$ where $\mathcal{C}$ is the set of integer labels. There are two types of losses commonly used for metric learning: embedding losses~\cite{hadsell2006dimensionality,weinberger2009distance,oord2018representation} that compare different instances (e.g., triplet loss), and proxy-based losses~\cite{wang2017normface,liu2017sphereface,zhai2018classification}
which learn parameters representing categories (e.g. weight vectors in normalized softmax loss or ProxyNCA~\cite{movshovitz2017no}).
As the embedding losses usually require specific sampling strategies to work well, we follow the latter for metric learning in this work. 

Specifically, for each category we define a learnable category center vector (or category embedding) ${\{\boldsymbol{c}_k\}}_{k=1}^c$ and optimize a distance metric so that instances of the same category are close and instances of different categories are far apart, as illustrated in the green shaded region of Fig.~\ref{fig:overview} and Fig.~\ref{fig:teasor}(b).  We optimize the following loss between learned shape instance embeddings $\boldsymbol{s}$ and category center embeddings $\boldsymbol{c}$:
\begin{equation}
    \mathcal{L}_{metric} = -\sum_{i=0}^N log \frac{exp(d(\boldsymbol{s}_{i}, \boldsymbol{c}_{y_i})/\tau)}{\sum_{k \in \mathcal{C}} exp(d(\boldsymbol{s}_{i}, \boldsymbol{c}_{k})/\tau)},
\end{equation}
where $d$ is a similarity measure and $\tau$ is the temperature. Following normalized softmax loss~\cite{wang2017normface} we use cosine similarity as a similarity measure,
\begin{equation}
    d(\boldsymbol{s}, \boldsymbol{c}) = \frac{\boldsymbol{s}\cdot \boldsymbol{c}}{\|\boldsymbol{s}\| \|\boldsymbol{c}\|}.
\end{equation}
The centers are randomly initialized from a uniform distribution and updated directly through gradient descent. We set temperature $\tau$ to be 0.3 across all our experiments.

\subsection{Shape and viewpoint regularization}
\label{subsection:regularization}

\vspace{-1mm} \noindent \textbf{Adversarial regularization.}
To further facilitate learning, we use adversarial regularization to ensure that predicted shapes and texture fields will render realistic images from any viewpoints in addition to the input view. Given the training image collection, we estimate the data manifold via an image discriminator. If the rendered images also lie on the training data manifold, most erroneous shapes and texture predictions will be eliminated. To achieve this, similar to~\cite{henzler2019escaping,ye2021shelf}, we use a discriminator with the adversarial training~\cite{goodfellow2014generative}. Furthermore, because our multi-category setting results in a complex image distribution, we condition our discriminator on category labels for easier training following~\cite{miyato2018cgans}. 
While our work is not the first work to use GANs for shape learning in general, we believe it still provides useful new insights by combining adversarial training with unconstrained implicit representation and class conditioning under our challenging MCSV setting.


Formally, suppose $I_{recon} = \boldsymbol{R}(\boldsymbol{f_S}, \boldsymbol{f_T}, \boldsymbol{v})$ is the output of the renderer for estimated shape, texture and viewpoint given an input image. 
We sample another random viewpoint $\boldsymbol{v}^\prime$ from a prior distribution $P_v(\boldsymbol{v})$ and render another image,
$I_{rnd} = R(\boldsymbol{f_S}, \boldsymbol{f_T}, \boldsymbol{v}^\prime)$ from this viewpoint.
We match the distribution of ${I}$ and ${I_{recon}, I_{rnd}}$ with adversarial learning as shown in Fig.~\ref{fig:overview}. Specifically, we optimize the objective $\mathcal{L}_{gan}$ below by alternatively updating our model (the $\boldsymbol{E}, \boldsymbol{H}, \boldsymbol{R}, \boldsymbol{V}$ networks) and the discriminator $\boldsymbol{D}(I;\boldsymbol{\theta_D})$
\begin{multline}
  \min_{\boldsymbol{\theta_E},\boldsymbol{\theta_H},\boldsymbol{\theta_V},\boldsymbol{\theta_R}} \max_{\boldsymbol{\theta_D}} \mathcal{L}_{gan} = \mathbb{E}[\log \boldsymbol{D}(I)] + \mathbb{E}[\log (1-\boldsymbol{D}([I_{recon},I_{rnd}]))].
\end{multline}

Here, $[I_{recon},I_{rnd}]$ is the stack of reconstructed and randomly rendered images (on batch dimension). For the update step of the reconstruction model, we follow a non-saturating scheme~\cite{goodfellow2014generative} where we maximize $\mathbb{E}[\log (\boldsymbol{D}([I_{recon},I_{rnd}]))]$ instead of minimizing $\mathbb{E}[\log (1-\boldsymbol{D}([I_{recon},I_{rnd}]))]$. We also use $R_1$ regularizer~\cite{mescheder2018training} and spectral normalization~\cite{miyato2018spectral} to stabilize the training process.

\vspace{1mm} \noindent \textbf{Viewpoint regularization via cycle-consistency.}
We regularize the viewpoint prediction module via cycle-consistency~\cite{zhu2017unpaired}, with a similar approach as the state-of-the-art at self-supervised viewpoint estimation~\cite{mustikovela2020self} and Ye et al.~\cite{ye2021shelf}. However,
they both rely on a strong shape symmetry assumptions to facilitate the learning and Ye et al. further utilize shape embedding in the cycle. In our approach, we do not use any such restrictive assumptions, and we only consider viewpoint itself in this regularization.

\begin{wrapfigure}{r}{0.45\linewidth}
\vspace{-35pt}
    \begin{center}
        \includegraphics[width=1\linewidth]{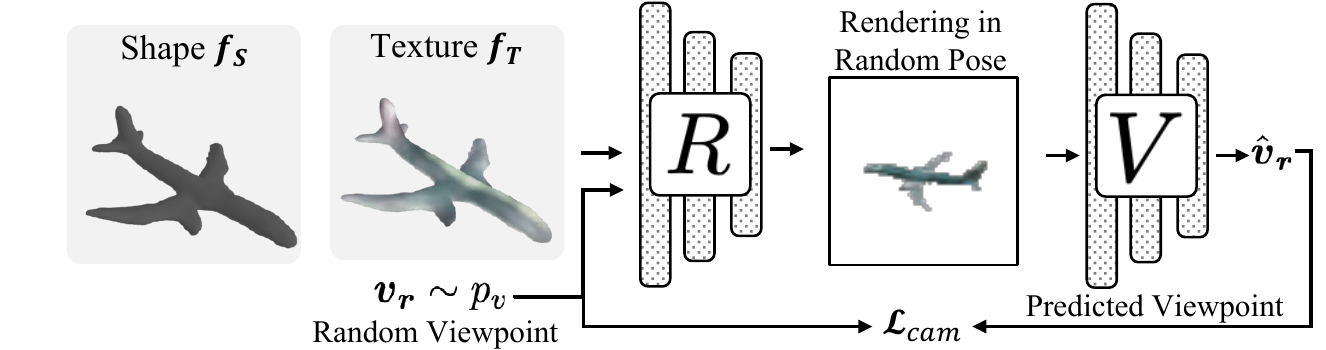}
    \end{center}
    \caption{Viewpoint regularization of our method. With a given shape and texture field, we sample a random viewpoint and render an image as the input to our viewpoint predictor. The viewpoint predictor is then supervised with the randomly sampled viewpoint, which forms a consistency cycle.}
    \label{fig:camera}
\vspace{-15pt}
\end{wrapfigure}

The key idea of viewpoint regularization is that given an image, we require that the viewpoint predictor accurately predicts the viewpoint of this image. Different from real images, we can render an arbitrary number of images by sampling viewpoints with a given shape and texture. This can be thought of as creating pseudo data-label pairs for the viewpoint predictor. 

Formally, given shape $\boldsymbol{f_S}$, texture field $\boldsymbol{f_T}$ and a random viewpoint $\boldsymbol{v_r}$, we render an image $\hat{I} = \boldsymbol{R}(\boldsymbol{f_S}, \boldsymbol{f_T}, \boldsymbol{v_r})$. The goal is to minimize the distance between $\boldsymbol{v_r}$ and $\hat{\boldsymbol{v_r}}= \boldsymbol{V}(\hat{I})$. With the trigonometric function representation, we maximize the cosine similarity between $\boldsymbol{v_r}$ and $\hat{\boldsymbol{v_r}}$ by minimizing
\begin{equation}
    \mathcal{L}_{cam} = 1 - \langle \boldsymbol{v_r}\,, \hat{\boldsymbol{v_r}}\rangle = 1 - \langle \boldsymbol{v_r}\,, \boldsymbol{V}(\boldsymbol{R}(\boldsymbol{f_S}, \boldsymbol{f_T}, \boldsymbol{v_r}))\rangle,
\end{equation}
where $\langle \cdot \,, \cdot \rangle$ denotes dot product. In practice, we apply this regularization on the reconstructed image $I_{recon}$ and the randomly rendered image $I_{rnd}$, together with the viewpoints that render them. This is computationally efficient as both images are also used in other losses. We use $\mathcal{L}_{cam}$ only to train the viewpoint prediction module, as we stop the gradients to shape, texture and rendering modules.
\section{Experiments}
\label{section:exp}
This section presents the empirical findings of applying our method to multiple synthetic and real datasets, as well as ablations of its individual components. An additional goal is to quantify the value of category labels in the MCSV setting. By comparing category label-based supervision with a standard two-view reconstruction approach, we find that the value of the category label is approximately equal to 20\% of an additional view. See the appendix for details.
Following an overview of the datasets, metrics and baselines, we give results on synthetic and real data. 

\subsection{Datasets} For synthetic data we use two splits of the ShapeNet~\cite{chang2015shapenet} dataset, and for real data we use Pascal3D+\cite{xiang2014beyond}.

\vspace{1mm} \noindent \textbf{ShapeNet-13.} This dataset consists of images from the 13 biggest categories of ShapeNet, originally used by Kato et al.~\cite{kato2018neural} and following works learning 3D shape without explicit 3D supervision~\cite{niemeyer2020differentiable,lin2020sdf}. 24 views per object are generated by placing cameras at a fixed $30^{\circ}$ elevation and with azimuthal increments around the object of $15^{\circ}$. Following SDF-SRN~\cite{lin2020sdf}, we use a 70/10/20 split for training, validation and testing, resulting in 30643, 4378 and 8762 objects respectively. We use only one view per object instance, which is randomly pre-selected out of the 24 views for each object.

\vspace{1mm} \noindent \textbf{ShapeNet-55.} To create a more challenging setting in comparison to prior work, we use all of the categories of ShapeNet.v2. This is challenging due to (1) approximately four times more shape categories; and (2) uniformly sampling at random the azimuth in the $[0^{\circ},360^{\circ}]$ range and elevation in $[20^{\circ},40^{\circ}]$. As a result, the image and shape distributions that the model needs to learn are significantly more complex. To reduce the data imbalance, we randomly sample at most 500 objects per category. Like ShapeNet13 we use
only one image per object instance and a 70/10/20 split, with 15031, 1705, 3427 images respectively.

\begin{figure*}[t]
\centering
	\vspace{-10pt}
	\includegraphics[width=0.9\linewidth]{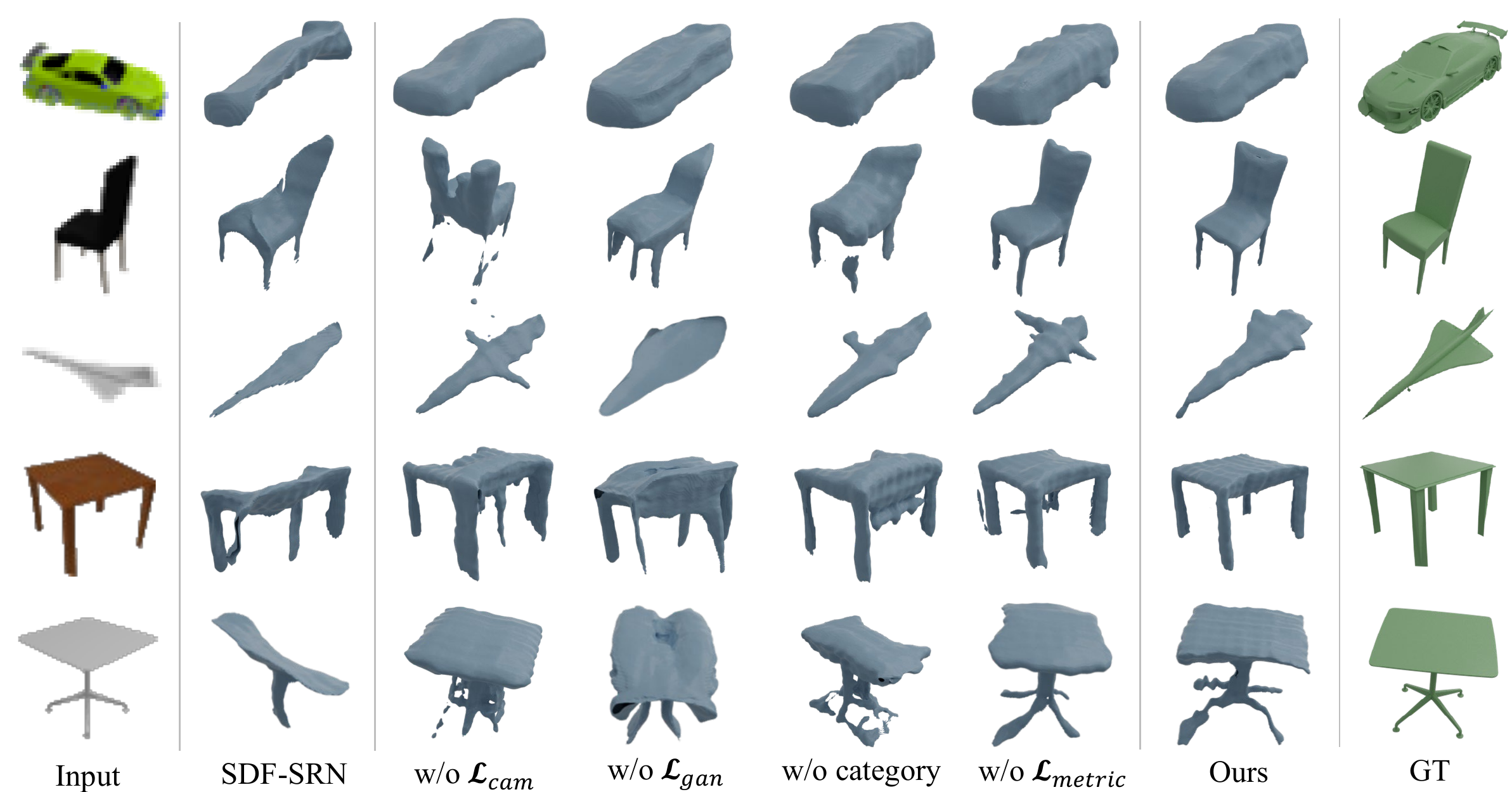}
	\caption{Qualitative comparison on ShapeNet-13. Our method learns both better global structures and details on various categories.}
	\label{fig:shapenet13_qual}
	\vspace{-20pt}
\end{figure*}

\vspace{1mm} \noindent \textbf{Pascal 3D+.} This dataset contains real-world images with 3D ground truth obtained by CAD model alignment.
Its challenges compared to ShapeNet come from (1) inaccurate object masks that include noisy backgrounds---an especially difficult setting for adversarial learning; and (2) the viewpoints vary in azimuth, elevation and tilt, creating challenges because different categories have different and unknown viewpoint distributions. We combine the commonly used motorcycle and chair categories for a total of 2307 images, and as in~\cite{lin2020sdf} we use the ImageNet~\cite{deng2009imagenet} subset of Pascal3D+ with 1132/1175 images for training/testing.

\vspace{1mm} \noindent \textbf{Evaluation Metrics}
We use Chamfer Distance (CD) and F-score under different thresholds to evaluate shape reconstruction performance following~\cite{lin2020sdf,groueix2018,tatarchenko2019single,thai20203d}. We use the Marching Cubes algorithm~\cite{lorensen1987marching} to convert the implicit representation to meshes prior to computing the metrics. Shapes are transformed into camera coordinates for evaluation, see the appendix for details.


\vspace{1mm} \noindent \textbf{Chamfer Distance.}
Following~\cite{niemeyer2020differentiable,lin2020sdf}, CD is defined as an average of accuracy and completeness. Given two surface point sets $S_1$ and $S_2$, CD is given as:
\begin{equation}
    d_{CD}(S_1, S_2) = \frac{1}{2|S_1|}\sum_{x \in S_1} \min_{y \in S_2} \|x-y\|_2 + \frac{1}{2|S_2|}\sum_{y \in S_2} \min_{x \in S_1} \|x-y\|_2
\end{equation}
\vspace{-15pt}

\vspace{1mm} \noindent \textbf{F-score.}
Compared to CD which accumulates distances, F-score measures  accuracy and completeness by thresholding the distances. For a threshold $d$ (F-Score@$d$), precision is the percentage of predicted points that have neighboring ground truth points within distance $d$, recall is the percentage of ground truth points that have neighboring predicted points within $d$. F-score, the harmonic mean of precision and recall, can be intuitively interpreted as the percentage of surface that is correctly reconstructed. 

\subsection{Baselines} There are no directly comparable state-of-the-art methods under our MCSV setting. Instead, we slightly alter the setting of two recent methods SDF-SRN~\cite{lin2020sdf} and Ye et al.~\cite{ye2021shelf} to compare with our model.

\vspace{1mm} \noindent \textbf{SDF-SRN}~\cite{lin2020sdf} learns to predict implicit shapes using single images and known camera poses in training. 
For comparison purposes, we attach our viewpoint prediction module to SDF-SRN and train it across multiple categories simultaneously. We compare to SDF-SRN on all datasets.

\vspace{1mm} \noindent \textbf{Ye et al.}~\cite{ye2021shelf} learns category-specific voxel prediction without viewpoint GT.\footnote{Their method has an optional per-sample test-time optimization to further refine the voxels and convert into meshes. Using the authors' implementation of this optimization did not lead to improved results in our experiments, so we use the voxel prediction as the output for evaluation.} We compare our method with Ye et al. on Pascal3D+.\footnote{Evaluating it on ShapeNet is quite challenging as it requires training numerous category-specific models with shape alignment for each one.} 
For Ye et al., we train different models for different categories following their original setting and compute the average metric over categories. We use GT masks as supervision for all experiments.

\subsection{ShapeNet-13}
We perform experiments on ShapeNet-13 and show quantitative and qualitative results in Table~\ref{tbl:main-shapenet13} and Figure~\ref{fig:shapenet13_qual}.

\vspace{-30pt}
\begin{table}
\centering
\caption{Quantitative result measured by CD and F-score on ShapeNet-13. Our method performs favorably to baselines and other SOTA methods. }
\begin{tabular}{lccc|c}
\hline
Methods & F-Score@1.0$\uparrow$  & F-Score@5.0$\uparrow$ & F-Score@10.0$\uparrow$ & \ \ CD$\downarrow$\ \  \\ \hline
w/o category  & 0.1589 & 0.6261 & 0.8527 & 0.520                    \\ 
w/o $\mathcal{L}_{metric}$ & 0.1875 & 0.6864 & 0.8805 & 0.458                    \\ 
w/o $\mathcal{L}_{cam}$ & 0.1837 & 0.6741 & 0.8758 & 0.463                    \\ 
w/o $\mathcal{L}_{gan}$ & 0.1846 & 0.6437 & 0.8422 & 0.532                 \\ \hline
Ours                    & \textbf{0.2005} & \textbf{0.7168} & \textbf{0.8949} & \textbf{0.430}                   \\ 
SDF-SRN    & 0.1606 & 0.5441 & 0.7584 & 0.682                    \\  \hline
\end{tabular}
\label{tbl:main-shapenet13}
\end{table}
\vspace{-15pt}


                
\vspace{1mm} \noindent \textbf{Ablation Study.} We first analyze the results of ablating the multiple regularization techniques used in our approach.
In Table~\ref{tbl:main-shapenet13}, `w/o category' shows the results of the model 
without any category information and `w/o $\mathcal{L}_{metric}$' is the model with GAN conditioning only, by removing the metric learning component. Comparing them to our full model, we clearly see our metric learning helps utilize category information in a more efficient way (w/o $\mathcal{L}_{metric}$ vs Ours). We also see the benefit of having category information for shape reconstruction tasks (w/o category vs Ours). We further ablate on our camera (w/o $\mathcal{L}_{cam}$ vs Ours) and GAN regularization (w/o $\mathcal{L}_{gan}$ vs Ours) and quantify their value. Our qualitative result verifies these findings as well. We include more quantitative findings about the value of category labels for shape learning in the appendix.

\vspace{1mm} \noindent \textbf{SOTA Comparison.} Comparing with SDF-SRN in Table~\ref{tbl:main-shapenet13} we see that our approach improves on the adapted SDF-SRN by a 23.8\% (w/o category) and 37.0\% (with category, Ours) CD decrease. Qualitatively, our approach captures both the overall shape topology and details (legs of table in the bottom row) whereas SDF-SRN fails. These all demonstrated the effectiveness of our proposed model.


\subsection{ShapeNet-55}
\begin{figure}
\vspace{-15pt}
\centering 
\includegraphics[width=1\linewidth]{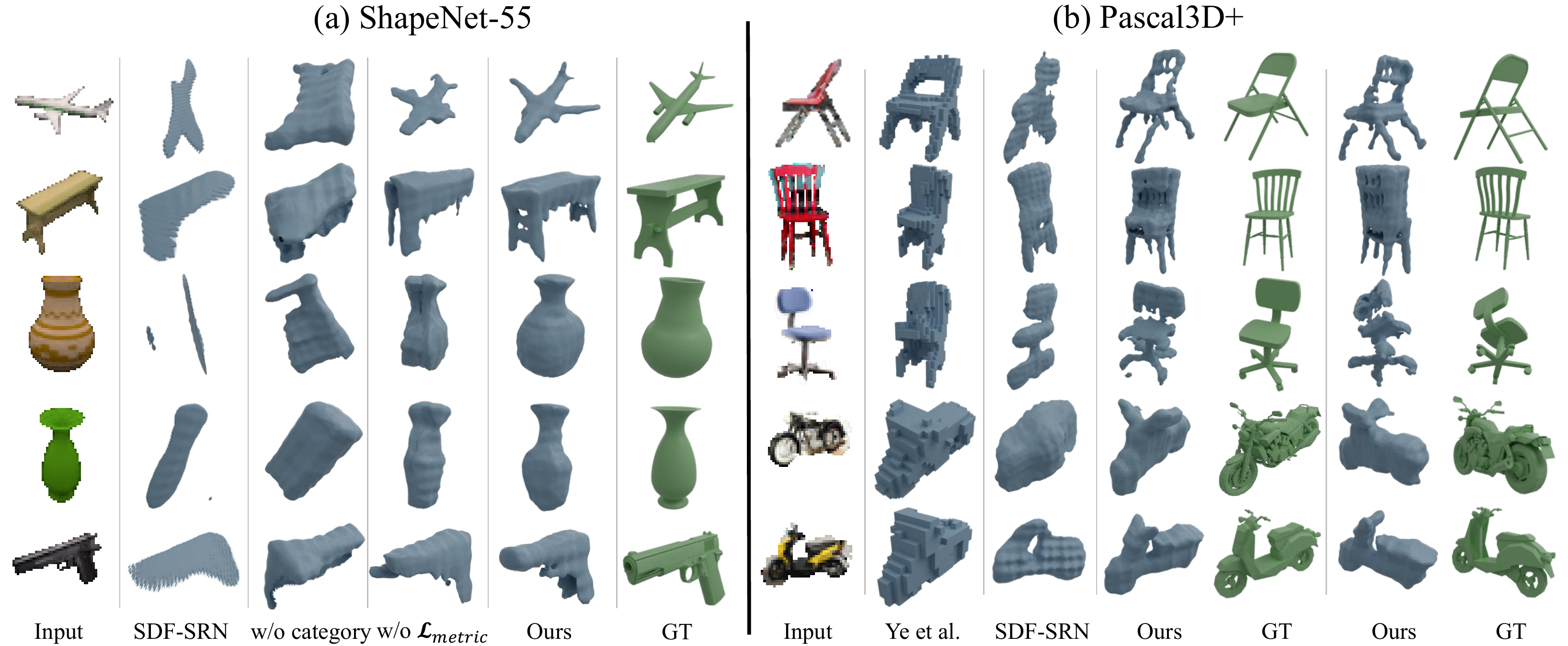}
\caption{Qualitative comparison on ShapeNet55 and Pascal3D+. Our method learns both better global 3D structure and shape details on various categories.}
\label{fig:shapenet55_pascal_qual}
\vspace{-15pt}
\end{figure}

Compared to ShapeNet-13, ShapeNet-55 is significantly more challenging due to both the number of categories and viewpoint variability. For results see Table~\ref{tbl:main-shapenet55} and Fig.~\ref{fig:shapenet55_pascal_qual} (a).

\begin{table}
\vspace{-25pt}
\centering
\caption{Quantitative result measured by CD and F-score on ShapeNet-55.  Our method performs favorably to baselines and other SOTA methods. }
\begin{tabular}{lccc|c}
\hline
Methods & F-Score@1.0$\uparrow$  & F-Score@5.0$\uparrow$ & F-Score@10.0$\uparrow$  & \ \ CD$\downarrow$ \ \  \\ \hline
w/o category  & 0.0977  & 0.4365 & 0.6815  & 0.801                    \\ 
w/o $\mathcal{L}_{metric}$  & 0.1431  & 0.5758 & 0.8047  & 0.620                    \\ 
Ours                          & \textbf{0.1619}  & \textbf{0.6164} & \textbf{0.8386}  & \textbf{0.541}                   \\ \hline
SDF-SRN~\cite{lin2020sdf}     & 0.0707  & 0.2750 & 0.4806  & 1.172                    \\  \hline
\end{tabular}
\label{tbl:main-shapenet55}
\vspace{-15pt}
\end{table}



\vspace{1mm} \noindent \textbf{Ablation Study.}
On this dataset, we again ablate on the value of category labels and our proposed shape metric learning. Similar to ShapeNet-13, we clearly see our metric learning method utilizes category information more effectively than GAN conditioning only (w/o $\mathcal{L}_{metric}$ vs Ours in Table~\ref{tbl:main-shapenet55}). Meanwhile, we demonstrate the huge benefit of having category information for shape reconstruction on this dataset (w/o category vs Ours, 32.4\% CD decrease). This is much more significant than ShapeNet-13 because ShapeNet-55 has much higher data complexity, where a better structured shape latent space can be quite beneficial to shape learning. Qualitatively, without category regularization the model cannot capture even the global shape successfully, whereas without shape metric learning it only partially succeeds at highly symmetric objects (two vases in Figure~\ref{fig:shapenet55_pascal_qual} (a)). For others (e.g. bench at row 2) its reconstruction can only explain the input view. This further strengthen the discussion in Fig.~\ref{fig:metric_comp_tsne} (a), where shape metric learning can help eliminate erroneous shape predictions.

\vspace{1mm} \noindent \textbf{SOTA Comparison.} Comparing with row 4 of Table~\ref{tbl:main-shapenet55} we see that our approach improves on the adapted SDF-SRN by 31.6\% (without category) and 53.8\% (with category) CD decrease. The significant performance improvement justifies the effectiveness of our overall framework, as well as the effectiveness of the categorical shape metric learning. Qualitatively, SDF-SRN collapses and fails to capture the topology of most inputs, while our method with still demonstrate a great qualitative reconstruction performance. More qualitative results on various categories and reconstructed textures are included in the appendix.

\subsection{Pascal3D+}


We compare our method to SDF-SRN~\cite{lin2020sdf} and Ye et al.~\cite{ye2021shelf} on Pascal3D+, as shown in Table~\ref{tbl:main-pascal} and Fig.~\ref{fig:shapenet55_pascal_qual} (b).
It is clear that our method outperforms the SOTA on both metrics, and our metric learning method is critical to the good performance. We also compare reconstructions qualitatively. As shown in Fig.~\ref{fig:shapenet55_pascal_qual} (b), SDF-SRN collapses to a thin flake that can only explain input images, while Ye et al. lack some important details. This further verifies the effectiveness of our proposed methods.



\begin{table}
\centering
\vspace{-25pt}
\caption{Quantitative result measured by CD and F-score on Pascal3D+.  Our method performs favorably to other SOTA methods. }
\begin{tabular}{lccc|c}
\hline
Methods  & F-Score@1.0$\uparrow$  & F-Score@5.0$\uparrow$ & F-Score@10.0$\uparrow$  & \ \ CD$\downarrow$\ \  \\ \hline
SDF-SRN~\cite{lin2020sdf}     & 0.0954  & 0.4656 & 0.7474 & 0.777                    \\  
Ye et al.~\cite{ye2021shelf}     & \textbf{0.1195}  & 0.5429 & 0.8252 & 0.625                    \\  
Ours w/o $\mathcal{L}_{metric}$     & 0.1038  & 0.5108 & 0.7799 & 0.673                 \\
Ours   & 0.1139  & \textbf{0.5460} & \textbf{0.8455}  & \textbf{0.580}                   \\ \hline
\end{tabular}
\label{tbl:main-pascal}
\vspace{-30pt}
\end{table}
\section{Limitation and Discussion}
\label{section:limitation}
One limitation is that reconstruction accuracy decreases for some samples with large concavity. For example, for bowls or bathtubs on ShapeNet-55, our method cannot capture the concavity of the inner surface. Concavity is hard to model without explicit guidance, and this can potentially be improved by explicitly modeling lighting and shading. On the other hand, ShapeNet-55 represents a class imbalance challenge, where images in different categories range from 40 to 500 images. This makes learning on rare classes difficult. 

We also observe a training instability of our model caused by the usage of adversarial regularization. Meanwhile, we think our shape metric learning can be further improved by explicitly modeling the multi-modal nature of some categories. This could be achieved by using several proxies for each category. We leave this to future work. Please see appendix for more discussion and qualitative examples on limitations.

\section{Conclusion}
\label{section:conclusion}
We have presented the first 3D shape reconstruction method with an SDF shape representation under the challenging \emph{multi-category, single-view (MCSV)} setting without viewpoint supervision. Our method leverages category labels to guide implicit shape learning via a novel metric learning approach and additional regularization. 
Our results on ShapeNet-13, ShapeNet-55 and Pascal3D+, demonstrate the superior quantitative and qualitative performance of our method over prior works. Our findings are the first to quantify the benefit of category information in single-image 3D reconstruction.

%
%
\bibliographystyle{splncs04}
\bibliography{refs}

\clearpage
\appendix
\section{Overview}
This appendix is structured as follows: In Section~\ref{sec:pix3d} we provide more experimental results on Pix3D~\cite{pix3d}; In Section~\ref{sec:quant-categ} we empirically quantify the value of using category labels as opposed to multi-view supervision; In Section~\ref{sec:shapenet-13-additional} we present additional SOTA comparison on ShapeNet-13; In Section~\ref{sec:implementation} we give implementation and training details for our model; In Section~\ref{sec:limitation} we discuss the limitations of our approach and in Section~\ref{sec:addnl-quali} we present additional qualitative results on our large-scale ShapeNet-55 renderings. 

\section{Additional Experiments on Pix3D}
\label{sec:pix3d}
\begin{figure}[h]
\centering
	\includegraphics[width=0.9\linewidth]{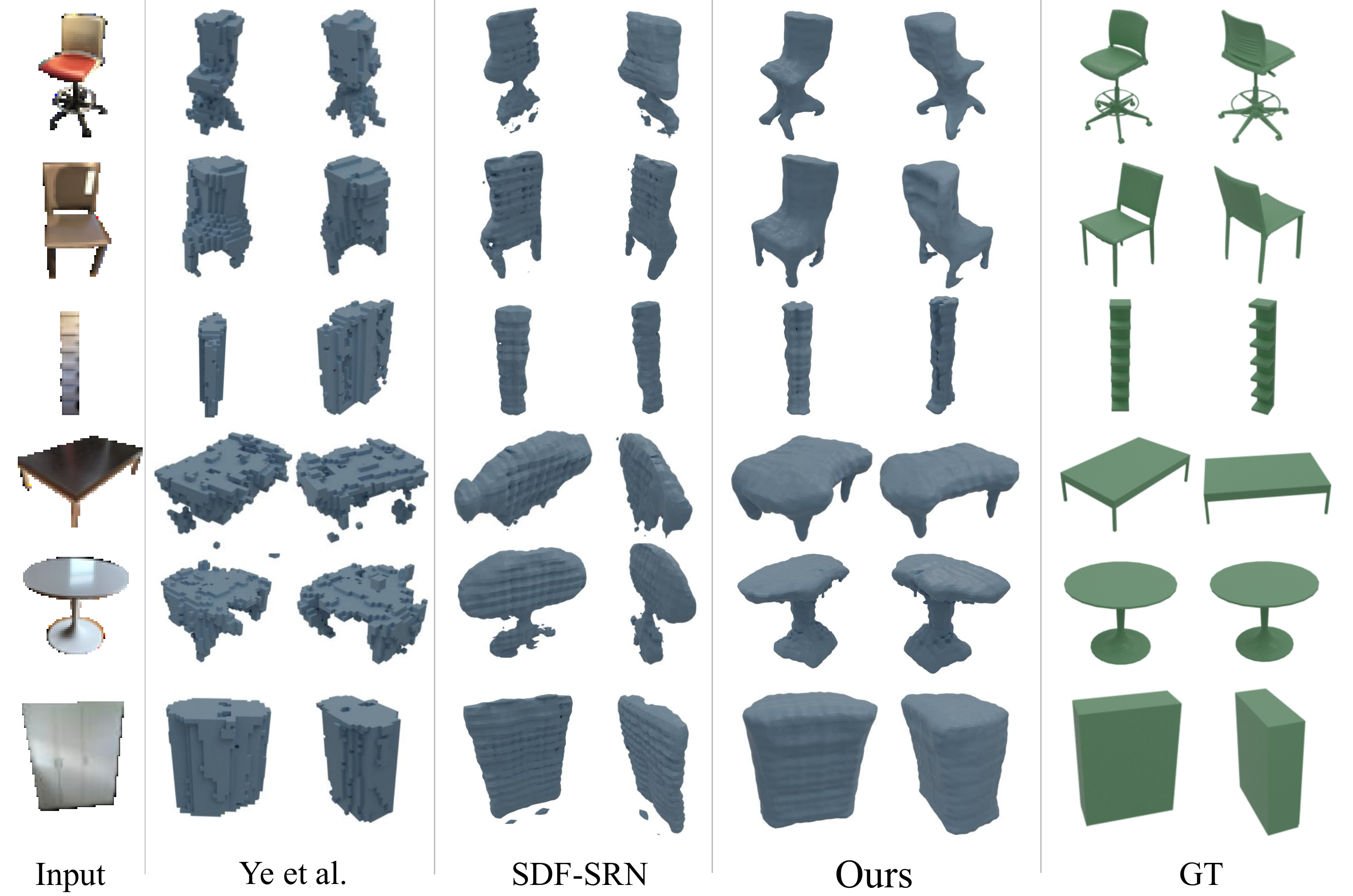}
	\caption{Qualitative comparison on Pix3D. Our method learns both better global 3D structure and shape details on various categories.}
	\label{fig:qual_pix3d}
	\vspace{-20pt}
\end{figure}

We additionally evaluate our methods on Pix3D~\cite{pix3d}. For Pix3D, we use 4 categories including bookcase, chair, table and wardrobe. We split the data with a 70/10/20 percentage into training, validation and testing similar to our experiments on ShapeNet. We compare our method to SDF-SRN~\cite{lin2020sdf} and Ye et al.~\cite{ye2021shelf} on Pix3D, as shown in Table~\ref{tbl:main-pix3d} and Fig.~\ref{fig:qual_pix3d}.
Again, it is clear that our method outperforms the SOTA methods quantitatively and qualitatively. These results further verify the effectiveness of our proposed methods. 

\begin{table}
\centering
\vspace{-20pt}
\caption{Quantitative result measured by CD and F-score on Pix3D.  Our method performs favorably to other SOTA methods. }
\begin{tabular}{lccc|c}
\hline
Methods  & F-Score@1.0$\uparrow$  & F-Score@5.0$\uparrow$ & F-Score@10.0$\uparrow$  & \ \ CD$\downarrow$\ \  \\ \hline
SDF-SRN~\cite{lin2020sdf}     & 0.1370 & 0.5622 & 0.7996 & 0.625                    \\  
Ye et al.~\cite{ye2021shelf}     & 0.1325 & 0.5308 & 0.7994 & 0.585                   \\  
Ours   		& \textbf{0.1745}  & \textbf{0.6604} & \textbf{0.8988}  & \textbf{0.421}                   \\ \hline
\end{tabular}
\label{tbl:main-pix3d}
\vspace{-10pt}
\end{table}

\section{Quantifying the value of category labels}
\label{sec:quant-categ}
\begin{figure}[h]
\centering
	\includegraphics[width=0.7\linewidth]{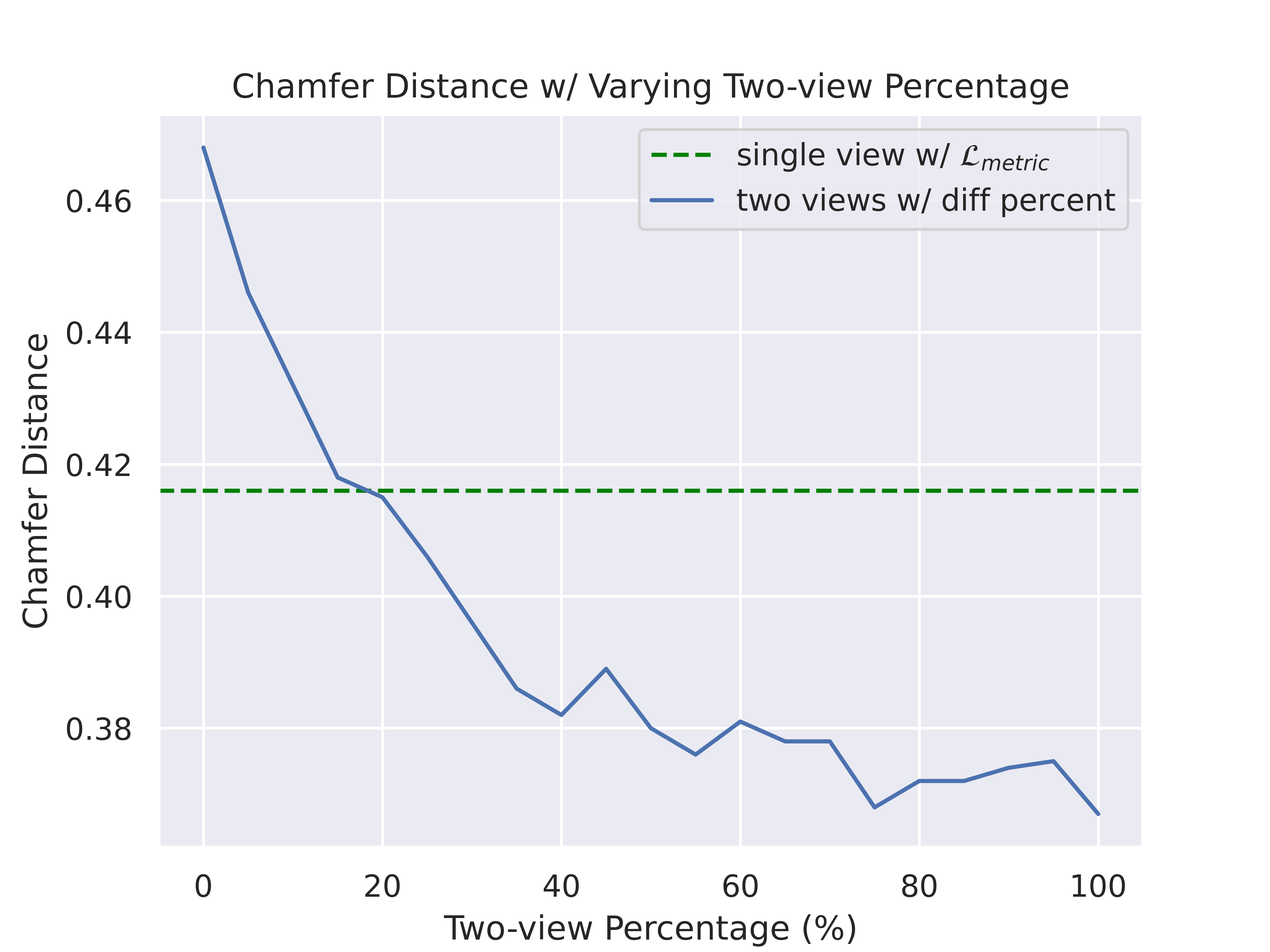}
	\caption{We evaluate the value of category label in MCSV reconstruction with camera pose. Under this quantitative evaluation, we show the category label will lead to similar performance of having access to around 15\% to 20\% two-view annotation.}
	\label{fig:cat-value}
\end{figure}

We further quantify the value of category label for the Multi-Category Single-View (MCSV) reconstruction task in this section. Our goal is to compare our method with category-guided shape metric learning with a model trained with multi-view supervision (more than a single view available per object instance). In this section, we train all models without adversarial regularization or the viewpoint predictor. We assume the viewpoint is known in this set of experiments. The baseline models are trained using an additional view as supervision. We follow a similar training process as~\cite{niemeyer2020differentiable} by randomly sampling a view per object for each epoch. To empirically determine the value of category labels, we vary the portion of the data that has multi-view annotations and compare that to our single-view category guided model. The quantitative results are shown in Figure~\ref{fig:cat-value}. As shown in the figure, having access to category labels can roughly lead to the reconstruction accuracy using 15\% to 20\% two-view annotation, measured by Chamfer Distance. Given the availability of category label compared to multi-view data, we believe that this is a promising finding. 

\section{Additional Comparison on ShapeNet-13}
\label{sec:shapenet-13-additional}
We additionally compare to Ye et al.~\cite{ye2021shelf} on ShapeNet-13. We train Ye et al. in their category-specific way (13 different models) and the comparison is shown in Table~\ref{tbl:comp-additional}. Our method outperforms Ye et al. significantly, even though the computation/parameters required for the category-specific models are much higher than ours.

\begin{table}
\centering
\vspace{-10pt}
\caption{Additional comparison measured by CD and F-score on ShapeNet-13.}
\begin{tabular}{lccc|c}
\hline
Methods  & F-Score@1.0$\uparrow$  & F-Score@5.0$\uparrow$ & F-Score@10.0$\uparrow$  & \ \ CD$\downarrow$\ \  \\ \hline
Ye et al.~\cite{ye2021shelf}     & 0.1349 & 0.5419 & 0.7777 & 0.669                   \\  
Ours  & \textbf{0.2005}  & \textbf{0.7168} & \textbf{0.8949}  & \textbf{0.430}                   \\ \hline
\end{tabular}
\label{tbl:comp-additional}
\vspace{-10pt}
\end{table}

\section{Implementation details}
\label{sec:implementation}
In this section, we provide more details about our architecture, loss functions, training and evaluation.

\subsection{Network architecture}
\label{sec:network}
Our architecture of image encoder, shape/texture module as well as the differentiable renderer share a similar design to SDF-SRN~\cite{lin2020sdf}, while shape and texture modules are made more lightweight for computational efficiency. The image encoder and the viewpoint predictor are based on ResNet18~\cite{he2016deep}. The hypernetwork that generates the weights of shape and texture field from latent code is a set of 6-layer MLPs of hidden dimension 512. Each MLP in this set generates the weights of a single layer of either shape or texture. The MLP representing $\boldsymbol{f_S}$ also has a 6-layer structure, with first 4 hidden layer 64 neurons and last hidden layer 32 neurons. The texture MLP has 2 layers, with a hidden dimension of 128. It is conditioned on the shape embedding following~\cite{yariv2020multiview}. Note that both shape and texture MLPs use Positional Encoding~\cite{mildenhall2020nerf} to encode input coordinates for better details. The differentiable renderer we use is from SDF-SRN~\cite{lin2020sdf}, where a LSTM~\cite{hochreiter1997long} learns to perform the ray marching steps. The LSTM predicts a step length for each rendering step based on local implicit feature as input and previous steps encoded as hidden state. We follow IDR~\cite{yariv2020multiview} (equation 7) to render the extra alpha channel, where we use the negative minimum SDF value on each ray with Sigmoid to represent the alpha value. In practice, we use the minimum SDF value from the ray-marcher steps instead of sampling numerous depths for each ray, and the SDF value is scaled by 30 before Sigmoid to increase the sharpness of the mask.

\subsection{Loss function} 
Our overall loss function is a weighted summation of the reconstruction loss, regularization losses and the losses that facilitate the learning of the renderer and the shape field as in SDF-SRN~\cite{lin2020sdf}:
\small
\begin{equation}
    \mathcal{L}_{total} = \mathcal{L}_{recon} + \lambda_1 \mathcal{L}_{metric} + \lambda_2 \mathcal{L}_{gan} + \lambda_3 \mathcal{L}_{cam} + \lambda_4 \mathcal{L}_{SDF-SRN}
\end{equation}
In our experiments, we set $\lambda_2=0.2$, $\lambda_3=0.03$, $\lambda_4=1$ for all datasets. We use a $\lambda_1$ of 0.1 for ShapeNet-55 and Pix3D, 0.05 for ShapeNet-13 and 0.03 for Pascal3D+. $\mathcal{L}_{render}$ itself is a weighted summation of several losses as well, we refer to~\cite{lin2020sdf} for more details. Since we render an extra alpha channel, we also use the commonly used soft IOU loss~\cite{liu2019soft} with coefficient 0.1 to supervise the predicted soft masks with GT masks. 

\subsection{Training and inference}
\label{sec:training}
\begin{figure}[h]
\centering
	\includegraphics[width=0.9\linewidth]{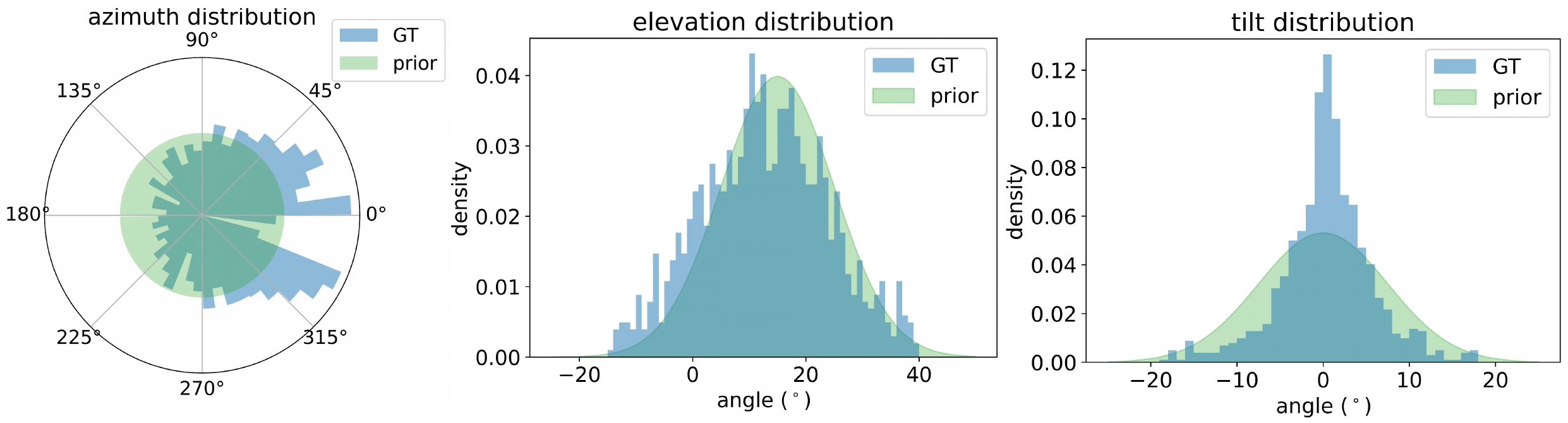}
	\caption{Comparison between our prior camera distribution (green) for training and the GT (blue) distributions on Pascal3D+.}
	\label{fig:camera_dist}
	\vspace{-15pt}
\end{figure}

To train our model, we iterate between the reconstruction step and the adversarial step. In the reconstruction step, the whole model except the discriminator is updated to minimize $\mathcal{L}_{total}$. All the regularizations we propose are activated during this step. In the adversarial step, only the discriminator is updated by maximizing $\lambda_1 \mathcal{L}_{gan}$, with all other loss disabled. For viewpoint sampling, we follow uniform distributions for azimuth, with a range of $[0^{\circ},360^{\circ}]$ across all datasets. The elevation is also sampled uniformly within $[20^{\circ},40^{\circ}]$ on ShapeNet-55. We sample elevation and tilt following Gaussian distributions on Pascal3D+ and treat mean and standard deviation as hyperparameters, similar to ~\cite{ye2021shelf}. Please see Fig.~\ref{fig:camera_dist} for the comparison of our prior distribution and the groundtruth camera distribution.

We use a learning rate of 0.0001 for both steps, and the optimizer is Adam~\cite{kingma2014adam} with $\beta_1 = 0$, $\beta_2=0.9$ and a batch size of 12. We did not use weight decay, learning rate scheduling or data augmentations. Our model is trained on a single NVIDIA GTX TITAN V for 80 epochs, which takes 2-3 days depending on the dataset size. The shape field is pretrained with the SDF values of a sphere for a better initialization as in SDF-SRN~\cite{lin2020sdf}. During inference, we only keep the image encoder and the shape prediction module. We implement our method in PyTorch~\cite{paszke2019pytorch} and will release the code upon publication. 

\subsection{Evaluation details}
\label{sec:evaluation}
We use the Marching Cubes algorithm~\cite{lorensen1987marching} to convert the implicit representation to meshes prior to computing the metrics. Specifically, we sample SDF values with a $128^3$ spatial grid and extract the 0-isosurface for marching cubes. We further sample 100000 points from each mesh for calculating the metrics. To align the predicted and GT shapes under the same canonical space, we transform both shapes to view-centered frames. 

Specifically, when training our models on Pascal3D+ and Pix3D, we assume weak-perspective cameras, and perform center crop and scaling over the input images. Since we do not know where the center of each GT shape is located w.r.t. the cropped and scaled image under a weak perspective camera, we align the meshes by registering shape predictions to the ground truth using the Iterative Closest Point (ICP) algorithm. This is in line with prior works such as SDF-SRN~\cite{lin2020sdf}.

\subsection{License}
\label{sec:license}
We develop our code based on the code of SDF-SRN~\footnote{https://github.com/chenhsuanlin/signed-distance-SRN} under the MIT license. We use ShapeNetV2~\cite{chang2015shapenet}, of which the license is specified at \url{https://shapenet.org/terms}. We use Pascal3D+~\cite{xiang2014beyond}
under a MIT license and Pix3D~\cite{pix3d} under a Creative Commons Attribution 4.0 International License.

\section{Limitations}
\label{sec:limitation}
We further discuss the limitations of our method in this section with more qualitative examples. When our model fails to reconstruct accurate shapes for some samples we observe it is primarily due to 3 reasons: 1) concavity, 2) class imbalance and 3) complex topology.

\begin{figure}[t!]
\begin{minipage}{0.48\textwidth}
\centering
	\includegraphics[width=\linewidth]{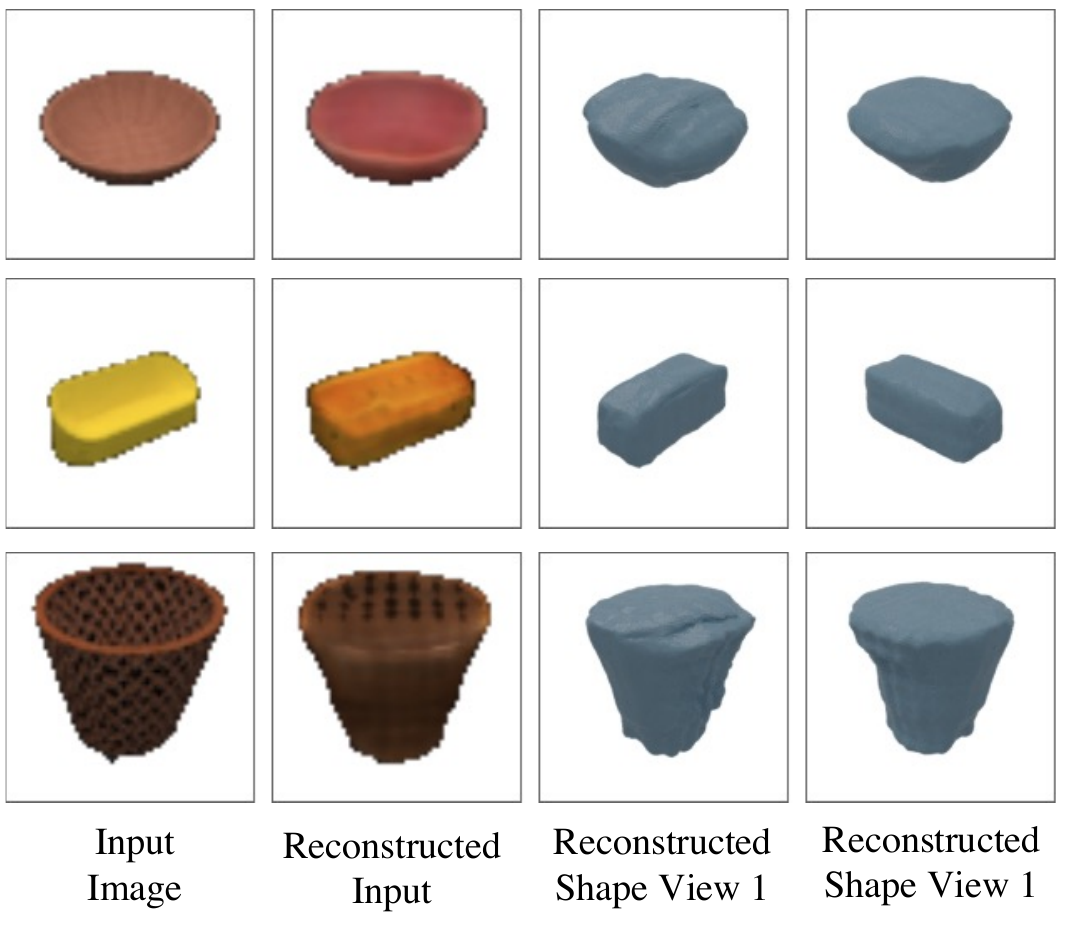}
	\caption{Illustration of limitation on concavity modeling. Our model fails to reconstruct concave regions for these samples.}
	\label{fig:concavity}
\end{minipage}
\hspace{7pt}
\begin{minipage}{0.48\textwidth}
\centering
	\includegraphics[width=\linewidth]{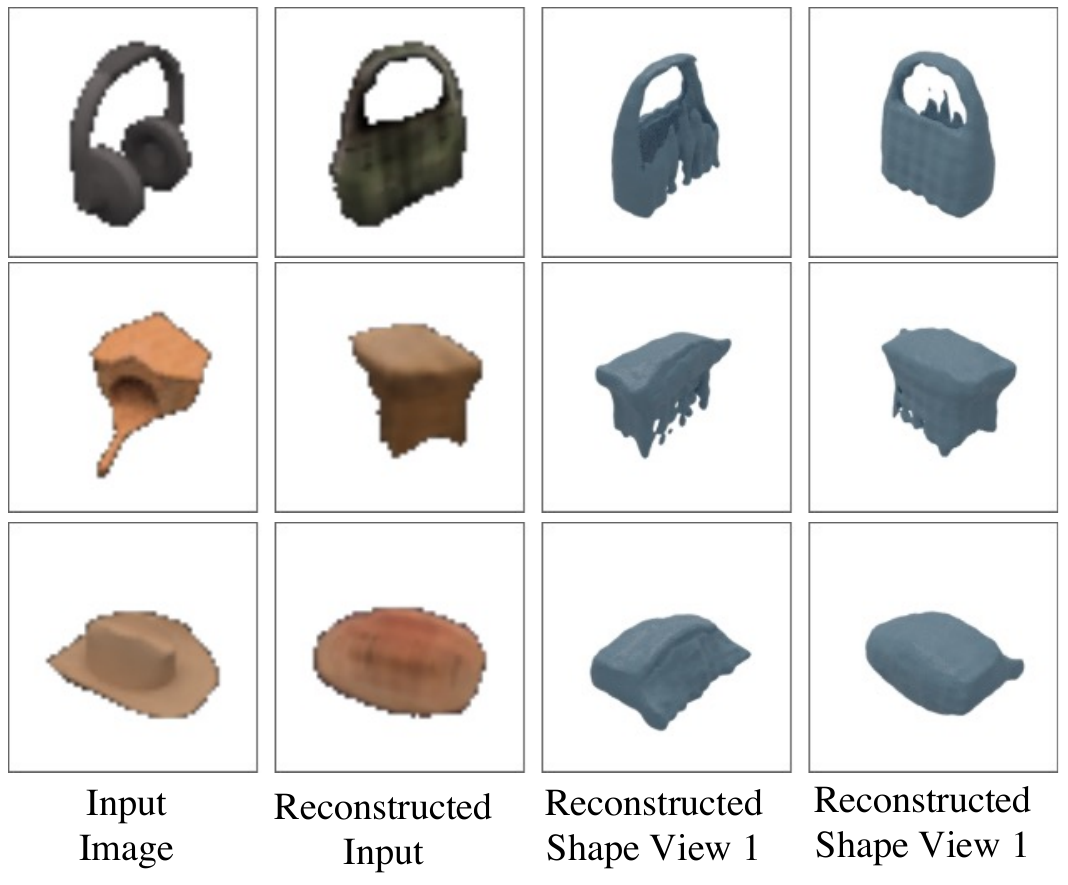}
	\caption{Illustration of limitation with class imbalance. Our model fails to reconstruct the shape of these rare categories accurately.}
	\label{fig:imbalance}
\end{minipage}
\vspace{-20pt}
\end{figure}

\vspace{1mm} \noindent \textbf{Concavity.} As discussed in the experiments, our method cannot model highly concave shapes such as bowls or hats. The masks for such shapes do not provide any information that reveals concavity. On the other hand, our method does not have access to explicit lighting or shading information. These factors make the learning of concavity hard. We demonstrate this issue in Fig.~\ref{fig:concavity} with examples from ShapeNet-55, including bowls, bath tubs and trash bins. We think it will be interesting to explore the explicit modeling of lighting/shading for future works.

\vspace{1mm} \noindent \textbf{Class imbalance.} We see a strong class imbalance on ShapeNet-55, where several classes have 500 samples while some only have 40 samples. Such an imbalance makes the learning challenging for some categories, as the gradient update within a minibatch can be dominated by major categories. We illustrate this issue in Fig.~\ref{fig:imbalance} by showing the reconstruction on 3 rare categories. We think it will be interesting to systematically explore the imbalance issue for shape reconstruction.


\begin{wrapfigure}{r}{0.5\linewidth}
\vspace{-20pt}
\centering
    \includegraphics[width=0.9\linewidth]{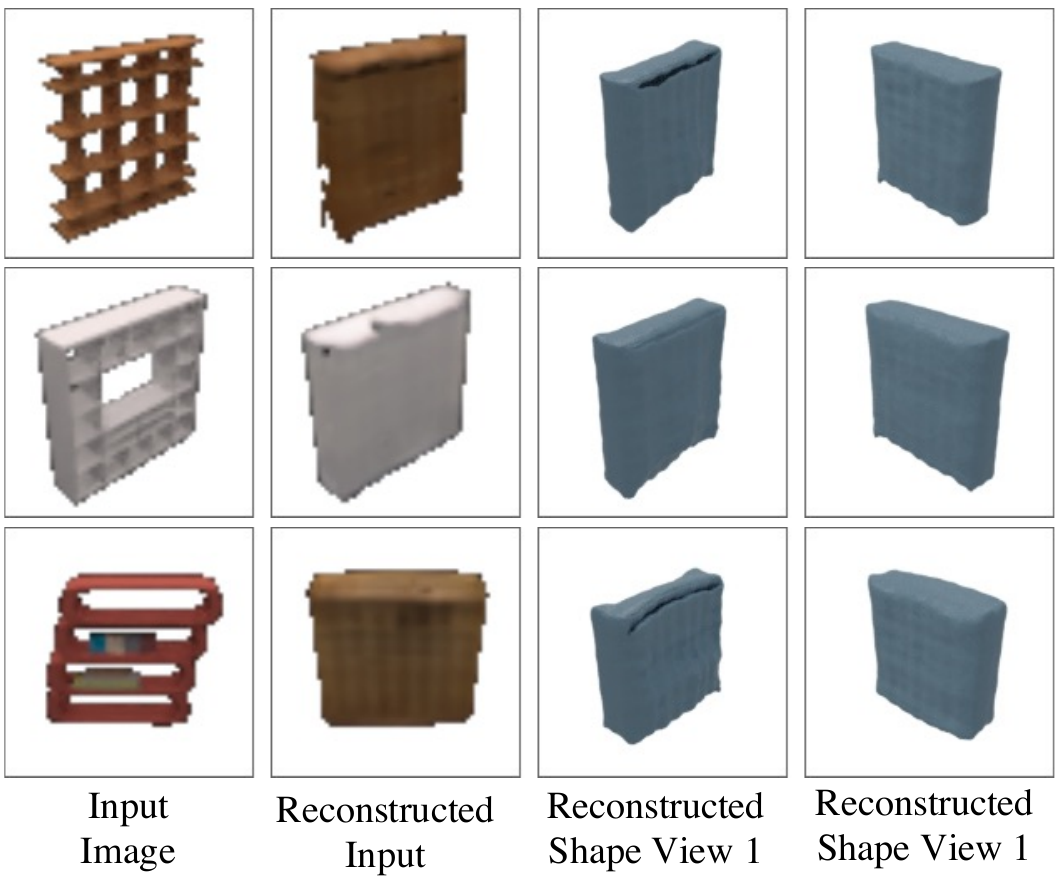}
	\caption{Illustration of complex topologies. For samples that have complex topologies, our model can only reconstruct a rough global structure.}
	\label{fig:topology}
\end{wrapfigure}

\vspace{1mm} \noindent \textbf{Complex topologies.} Due to the lack of 3D or multi-view supervision, it is still quite challenging to learn accurate shapes when the topologies are complex. We illustrate this issue in Fig.~\ref{fig:topology} by showing three shelves from ShapeNet-55. We believe it is still an open problem to learn accurate shapes for such examples under the challenging \emph{multi-category, single-view (MCSV)} setting without viewpoint supervision.

We hope these limitations are beneficial observations to inform and guide future research under similar challenging settings. On the other hand, despite these limitations, our method can reconstruct accurate shapes for the majority of images or categories. We believe this is a significant step toward fully unsupervised shape learning. 



\section{Additional Qualitative Results.}
\label{sec:addnl-quali}
\begin{figure*}[h]
\centering
	\includegraphics[width=1.0\linewidth]{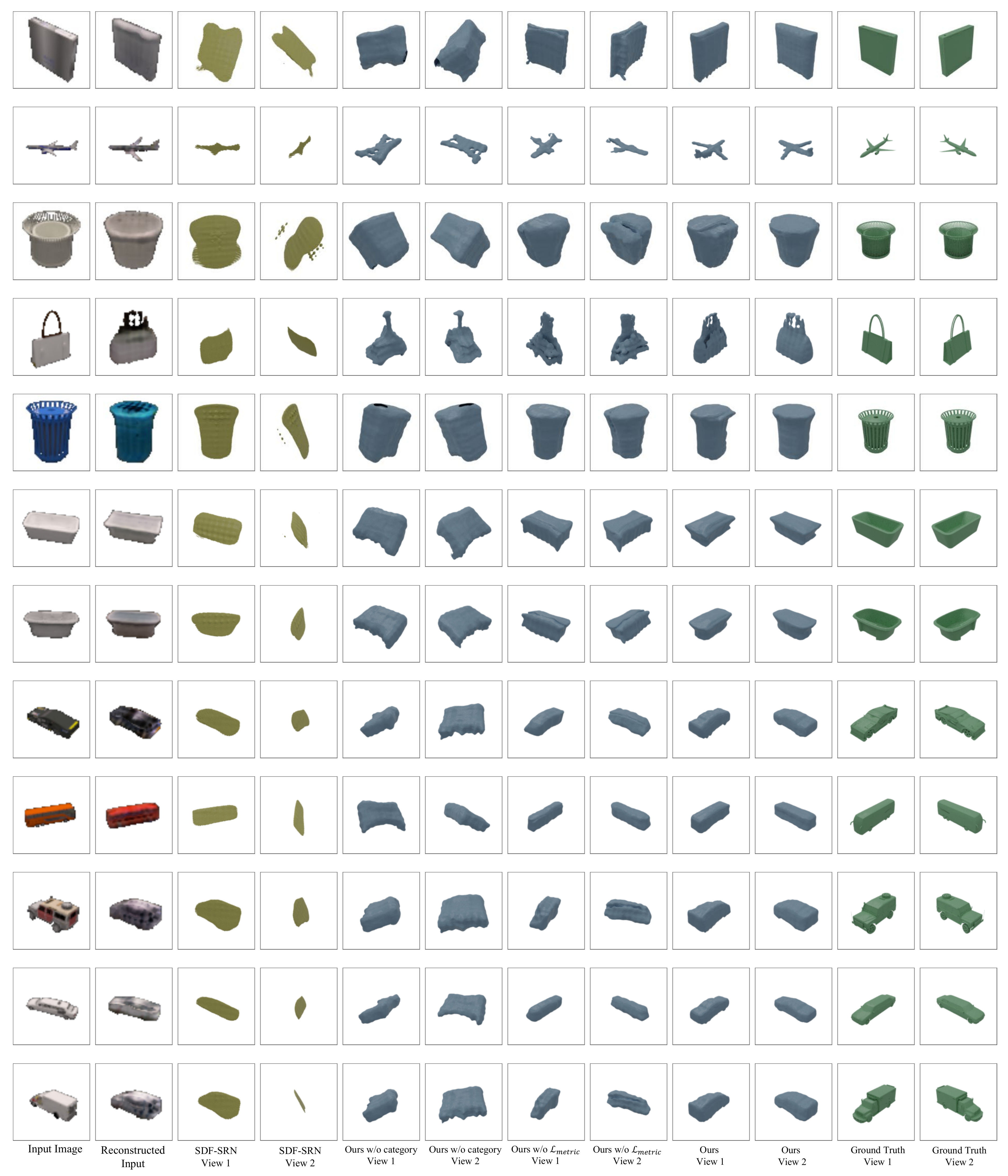}
	\caption{Additional qualitative results on ShapeNet-55.}
	\label{fig:shapenet55-1}
	\vspace{10mm}
\end{figure*}

\begin{figure*}[h]
\centering
	\includegraphics[width=1.0\linewidth]{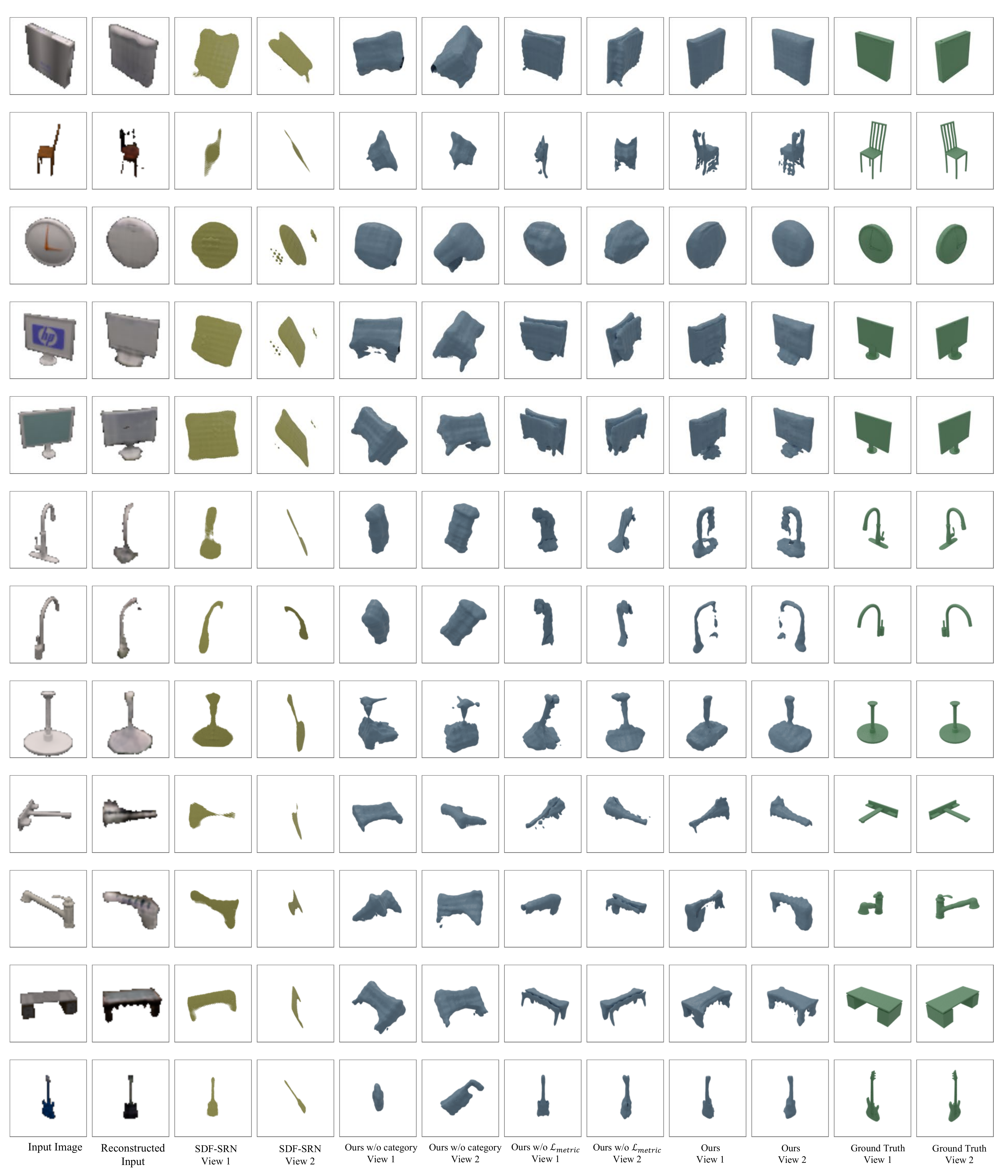}
	\caption{Additional qualitative results on ShapeNet-55.}
	\label{fig:shapenet55-2}
	\vspace{10mm}
\end{figure*}

\begin{figure*}[h]
\centering
	\includegraphics[width=1.0\linewidth]{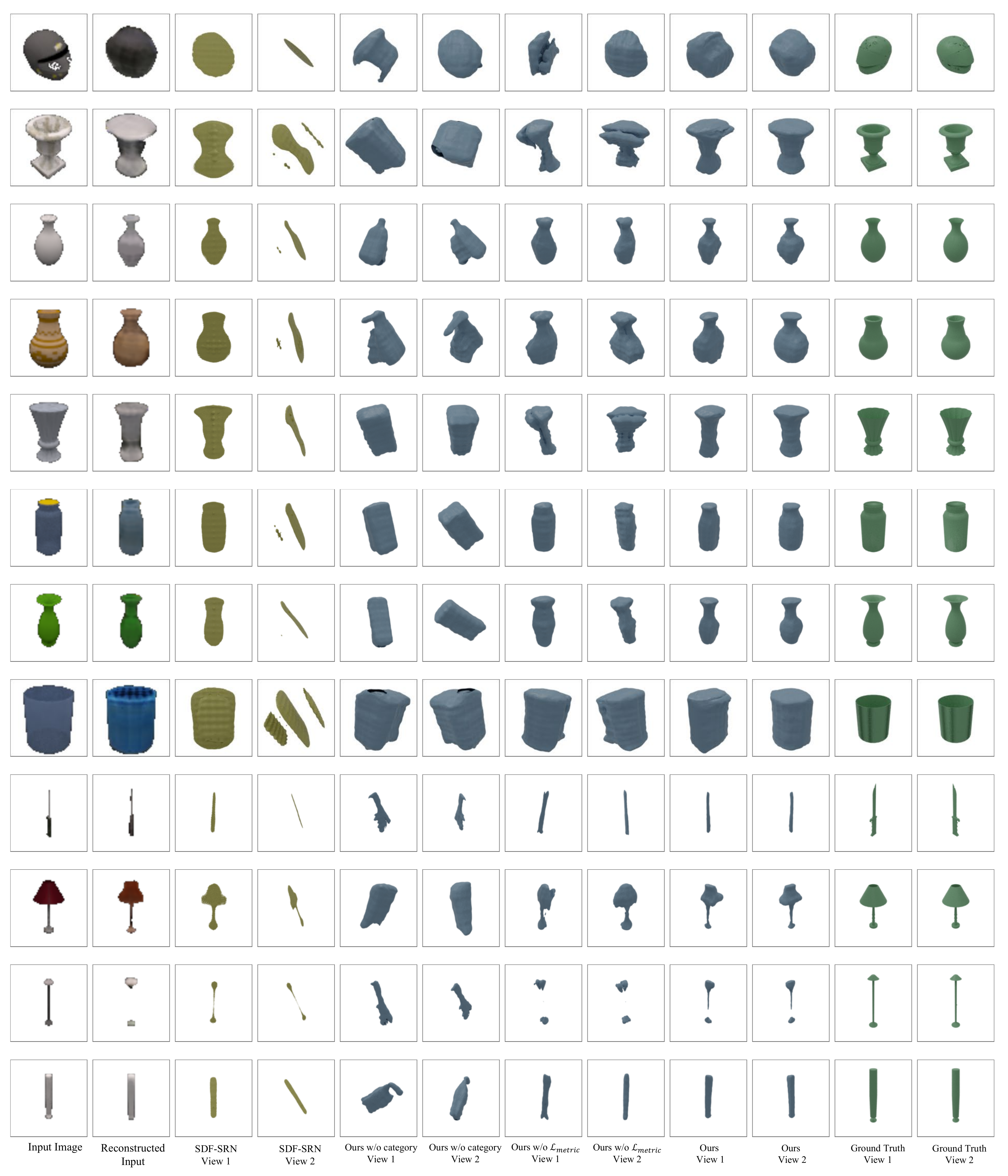}
	\caption{Additional qualitative results on ShapeNet-55.}
	\label{fig:shapenet55-3}
	\vspace{10mm}
\end{figure*}

\begin{figure*}[h]
\centering
	\includegraphics[width=1.0\linewidth]{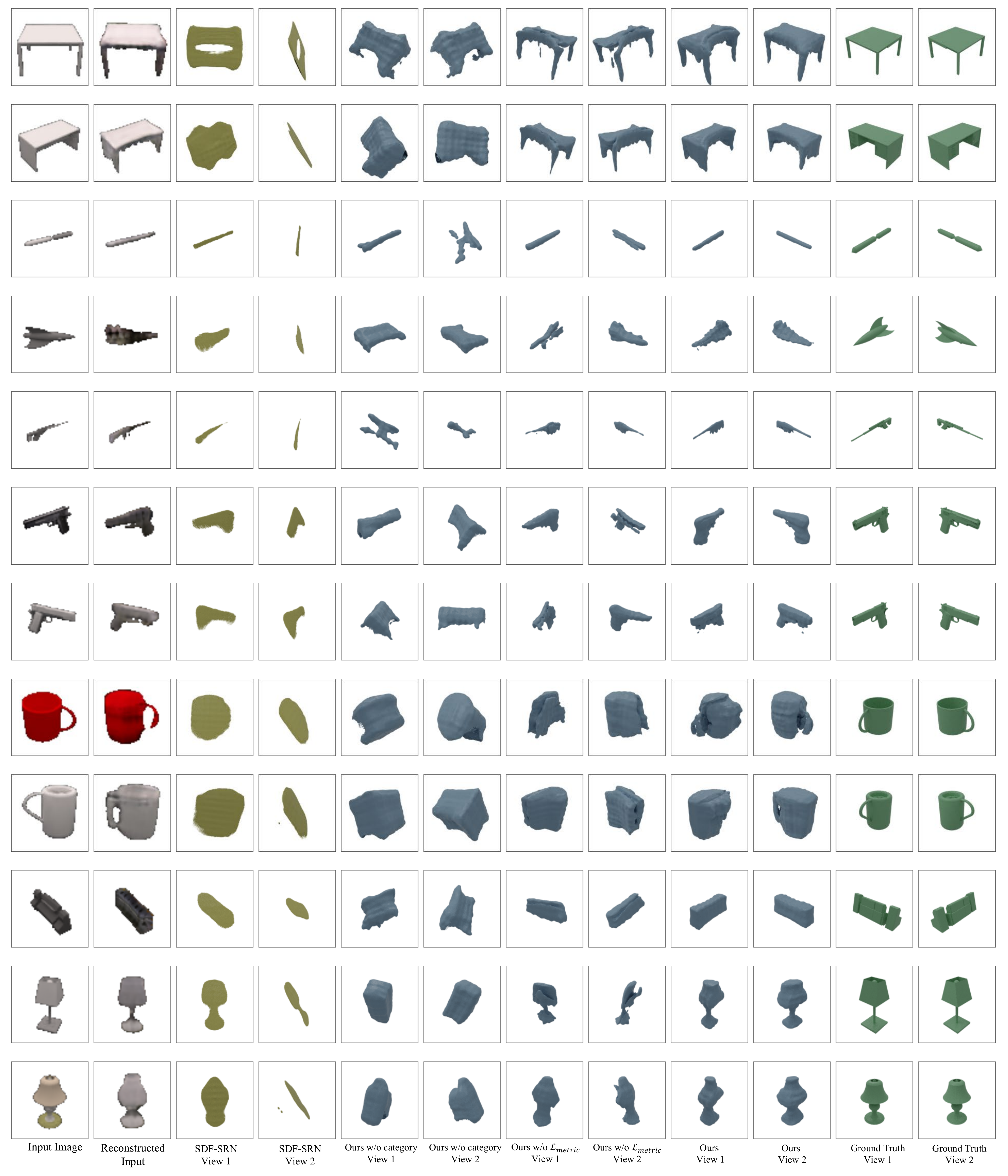}
	\caption{Additional qualitative results on ShapeNet-55.}
	\label{fig:shapenet55-4}
	\vspace{10mm}
\end{figure*}

In this section, we show more qualitative results of our model on ShapeNet-55 across various categories, as in Fig.~\ref{fig:shapenet55-1}, Fig.~\ref{fig:shapenet55-2}, Fig.~\ref{fig:shapenet55-3} and Fig.~\ref{fig:shapenet55-4}.

\end{document}